\documentclass[10pt,twocolumn,letterpaper]{article}

\usepackage[pagenumbers]{cvpr} 

\usepackage{graphicx}
\usepackage{amsmath}
\usepackage{amssymb}
\usepackage{booktabs}

\usepackage{algorithm}

\usepackage{amsmath}
\usepackage{amssymb}
\usepackage{makecell}
\usepackage{bbding}
\usepackage{pifont}
\usepackage{bigstrut}
\usepackage{color}
\usepackage{algorithm}
\usepackage{algpseudocode}
\usepackage{multirow}

\usepackage{colortbl}
\usepackage[table]{xcolor}
\definecolor{mygray}{gray}{.8}

\usepackage[pagebackref,breaklinks,colorlinks]{hyperref}

\usepackage[capitalize]{cleveref}
\crefname{section}{Sec.}{Secs.}
\Crefname{section}{Section}{Sections}
\Crefname{table}{Table}{Tables}
\crefname{table}{Tab.}{Tabs.}

\def\cvprPaperID{****} 
\def\confName{CVPR}
\def\confYear{2022}

\begin{document}

\title{Less is More: Learning from Synthetic Data with Fine-grained Attributes for Person Re-Identification}

\author{Suncheng Xiang$^{1}$, Guanjie You$^{2}$, Mengyuan Guan$^{1}$, Hao Chen$^{1}$, Binjie Yan$^{1}$, Ting Liu$^{1}$, Yuzhuo Fu$^{1}$\\
$^{1}$Shanghai Jiao Tong University, $^{2}$National University of Defense Technology\\
{\tt\small \{xiangsuncheng17, gemini.my, 958577057, yanbinjie, louisa\_liu, yzfu\}@sjtu.edu.cn, ygjssxz@163.com}
}


\twocolumn[{
\renewcommand\twocolumn[1][]{#1}
\maketitle
\begin{center}
	\centering
	\vspace{-1em}
	\includegraphics[width=\linewidth]{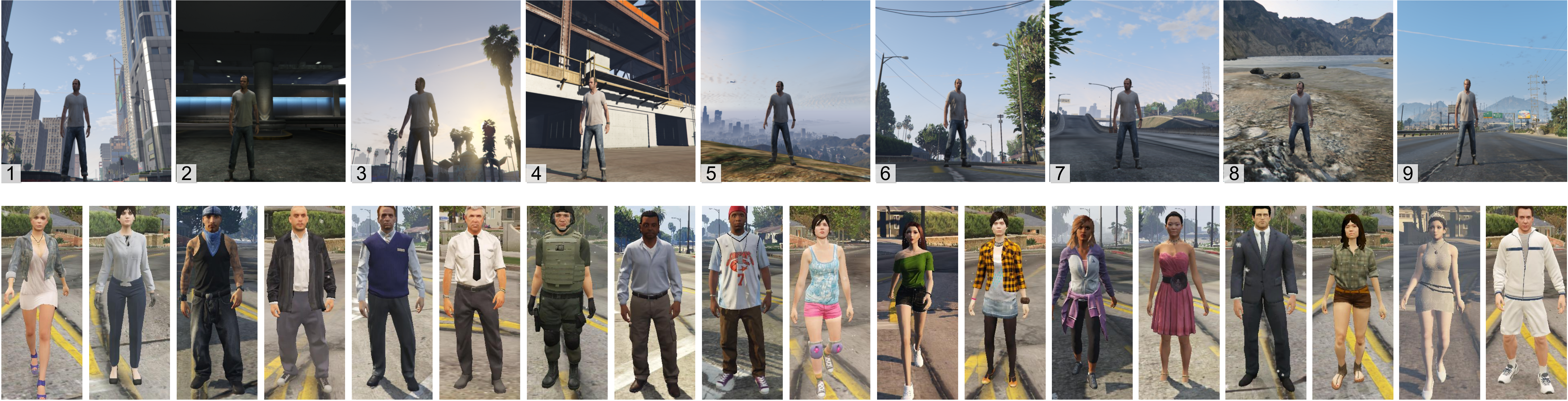}
	\vspace{-2pt}
	\captionof{figure}{Sample images from the proposed \emph{FineGPR} dataset, which contains 2,028,600 images of 1,150 identities. We manually labeled fine-grained attribute annotations both at environment level and identity level. \textit{First row}: With the same characters in different scenes, in each scene, a person can face toward a manually denoted direction. \textit{Second row}: Different characters in the same scene (\textit{i.e.} \textit{Scene \#6}).}
	\label{fig2}
	\vspace{-2pt}
\end{center}
}]
\maketitle

\begin{abstract}
      Person re-identification (re-ID) plays an important role in applications such as public security and video surveillance. Recently, learning from synthetic data, which benefits from the popularity of synthetic data engine, has attracted great attention from the public eyes. However, existing datasets are limited in quantity, diversity and realisticity, and cannot be efficiently used for re-ID problem. To address this challenge, we manually construct a large-scale person dataset named \textit{FineGPR} with fine-grained attribute annotations. Moreover, aiming to fully exploit the potential of \textit{FineGPR} and promote the efficient training from millions of synthetic data, we propose an attribute analysis pipeline called AOST, which dynamically learns attribute distribution
   in real domain, then eliminates the gap between synthetic and real-world data and thus is freely deployed to new scenarios. Experiments conducted on benchmarks demonstrate that \textit{FineGPR} with AOST outperforms (or is on par with) existing real and synthetic datasets, which suggests its feasibility for re-ID task and proves the proverbial \textit{\textbf{less-is-more}} principle.
   Our synthetic \textit{FineGPR} dataset is publicly available at \textit{\url{https://github.com/JeremyXSC/FineGPR}}.

\end{abstract}

\section{Introduction}
Given a query image, person re-identification aims to match images of the same person across non-overlapping camera views, which has attracted lots of interests and attention in both academia and industry. Encouraged by the remarkable success of deep learning networks~\cite{he2016deep,pan2018two} and the availability of re-ID datasets~\cite{ristani2016performance,zheng2015scalable}, performance of person re-ID has been significantly boosted and made great progress. However, in practice, manually labelling a large diversity of training data is time-consuming and labor-intensive when directly deploying re-ID system to new scenarios. During intensive annotation, one needs to associate a pedestrian across different cameras, which is a difficult and laborious process as people might exhibit very different appearances in different cameras. In addition, there also has been an increasing concern over data safety and ethical issues,
\textit{e.g.} DukeMTMC-reID~\cite{ristani2016performance,zheng2017unlabeled} has been taken down due to privacy problem.
Some European countries already passed privacy-protecting laws~\cite{goddard2017eu} to prohibit the acquisition of personal data
without authorization, which makes collection of large-scale datasets extremely difficult.

\begin{figure}[t]
\centerline{\includegraphics[width=0.95\linewidth]{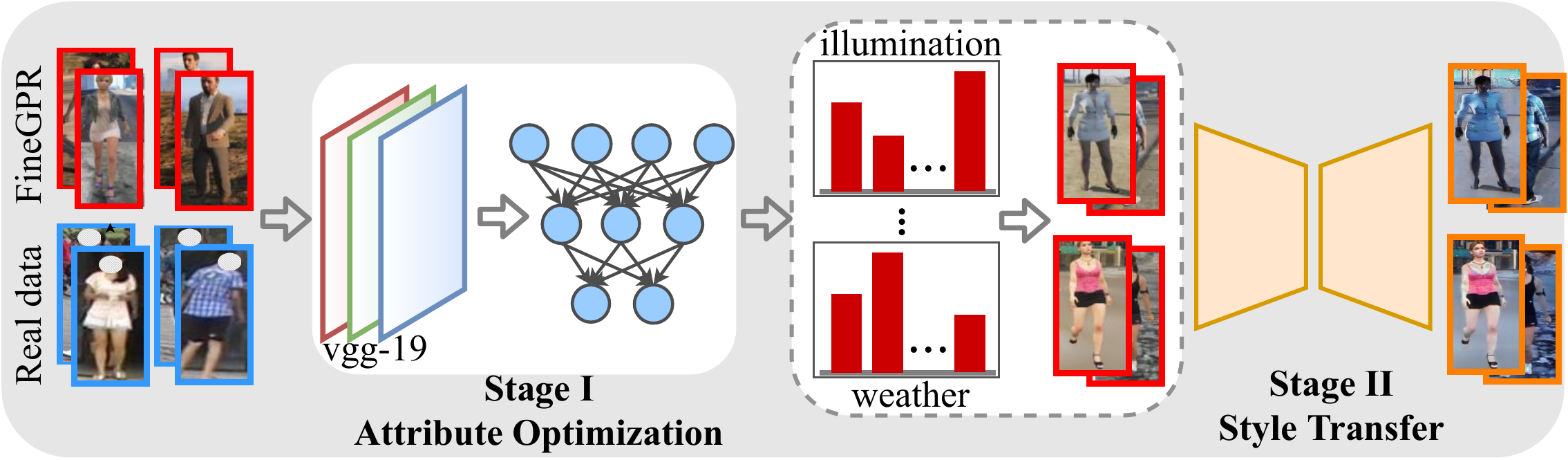}}
\caption{System workflow of the AOST method, which is based on the dataset with fine-grained attribute annotations.}
\label{fig1}
\end{figure}

To address this challenge, several works~\cite{bak2018domain,barbosa2018looking,sun2019dissecting,wang2020surpassing} have proposed to employ off-the-shelf game engines to generate synthetic person images. For example, Barbosa \textit{et al.}~\cite{barbosa2018looking} construct a SOMAset which contains 50 3D human models. SyRI~\cite{bak2018domain} provides 100 virtual humans rendered with 140 HDR environment maps.  Wang \textit{et al.}~\cite{wang2020surpassing} collect a RandPerson dataset with 1,801,816 synthesized person images.
Although these datasets provide considerable benefits of the data scale and enable some preliminary research in person re-ID, they are quite limited in both the attribute distribution and collected environment,
\textit{e.g.}, SyRI does not have concept of cameras, and SOMAset is uniformly distributed along an environment with clothing variations. In essence, these synthetic datasets either focus on independent attribute, or require annotators to carefully simulate specific scenes in detail, few datasets consider fine-grained attribute annotations or high-quality image resolution, which limits their scalability and diversity in terms of synthesized person. Another challenge we could observe is that, previous methods mainly focus on achieving competitive performance with large-scale data at the sacrifice of  expensive time costs and intensive human labors, while neglect to perform efficient training with a higher quality of attribute annotations from millions of synthetic data. Considering the fact that
existing real-world datasets can be very different in terms of content and style, \textit{e.g.}, Market-1501~\cite{zheng2015scalable} consists mostly of summer scenes captured in campus, while light in the CUHK03~\cite{li2014deepreid} covers a wide range of indoor scenes, directly using all synthetic dataset for training will undoubtedly produce negative effects for domain adaptation, which makes it infeasible in practical scenarios.

In order to alleviate the problems identified above and facilitate the study of re-ID community, we start from two perspectives, namely
data and methodology. From the aspect of  the data, we propose to collect data from synthetic world on the basis of GTA5 game engine, and
manually construct a \textbf{Fine}-grained \textbf{G}TA \textbf{P}erson \textbf{R}e-ID dataset called \textbf{\textit{FineGPR}}, which provides accurate and configurable annotations, \textit{e.g.}, \textit{viewpoint, weather, diverse and informative illumination and background}, as well as the various \textit{pedestrian attribute annotations at the identity level}. Compared to existing person re-ID datasets, \textit{FineGPR} is explicitly distinguished in richness, quality and diversity. It is worth noting that our data synthesis engine is still extendable to generate more data, which can be edited/extended not only for this study, but also for future research in re-ID community.

From the aspect of methodology, we introduce a novel \textbf{A}ttribute \textbf{O}ptimization and \textbf{S}tyle \textbf{T}ransfer pipeline \textbf{AOST} to perform training on re-ID task in a data-efficient manner. AOST can dynamically select samples which approximates the attribute distribution in real domain. As illustrated in Fig.~\ref{fig1},  the proposed AOST contains Stage-I (\textit{Attribute Optimization}) and Stage-II (\textit{Style Transfer}). Firstly,  Stage-I is adopted to mine the attribute distribution of real
domain, following by the Stage-II to reduce the intrinsic gap between synthetic
and real domain. Finally, the transferred data are adopted for performing training on downstream vision task.
This is the first time as far as we know, to greatly promote efficient training from millions of synthetic data on re-ID task,
experiments across diverse datasets suggest that the “\textit{\textbf{less-is-more}}” principle is highly effective in practice.

Our contributions can be summarized into three aspects:
\begin{itemize}
\item We open source the largest person dataset with fine-grained attribute annotations for the community without privacy and
ethics concerns.
\item  Based on it, we propose a two-stage pipeline AOST to learn from fine-grained attributes, then eliminate style differences between synthetic and real domain for more efficient training.
\item Extensive experiments conducted on benchmarks show that our \textit{FineGPR} is promising and can achieve competitive performance with AOST in re-ID task.
\end{itemize}

\begin{table*}[htbp]
  \centering
  \footnotesize
  \setlength{\tabcolsep}{0.70mm}{
    \begin{tabular}{l|l|c|c|c|c|c|c|c|c|c|c}
    \Xhline{0.8pt}
    \multicolumn{2}{c|}{Dataset} & \#Identities & \#Bboxes & \#Cameras & \#Wea. & \#Illum. & \#Scenes  &\#ID-level  & \#Resolution  & Hard samples  & Ethical Considerations  \\
    \hline\hline
    \multirow{4}[2]{*}{Real} & Market-1501~\cite{zheng2015scalable} & 1,501  & 32,668 & 6     & \textcolor{red}{\ding{56}}     & \textcolor{red}{\ding{56}}     & \textcolor{red}{\ding{56}}   &\textcolor[rgb]{0.00,0.50,0.00}{\ding{52}}    & low  & No   & No  \\
          & CUHK03~\cite{li2014deepreid} & 1,467  & 14,096 & 2     & \textcolor{red}{\ding{56}}     & \textcolor{red}{\ding{56}}     & \textcolor{red}{\ding{56}}   &\textcolor{red}{\ding{56}}    & low   & No  & No \\
          & MSMT17~\cite{wei2018person} & 4,101  & 126,441 & 15    & \textcolor{red}{\ding{56}}     & \textcolor{red}{\ding{56}}     & \textcolor{red}{\ding{56}}   & \textcolor{red}{\ding{56}}    & vary   & No   & No \\
    \hline
    \multirow{5}[4]{*}{Synthetic} & SOMAset~\cite{barbosa2018looking} & 50     & 100,000 & 250   & \textcolor{red}{\ding{56}}     & \textcolor{red}{\ding{56}}     & \textcolor{red}{\ding{56}}   & \textcolor{red}{\ding{56}}  & --   & No  & No \\
          & SyRI~\cite{bak2018domain}  & 100    & 1,680,000 & 280     & \textcolor{red}{\ding{56}}     & 140     & \textcolor{red}{\ding{56}}   & \textcolor{red}{\ding{56}}   & --   & No  & No \\
          & PersonX~\cite{sun2019dissecting} & 1,266   & 273,456 & 6     & \textcolor{red}{\ding{56}}     & \textcolor{red}{\ding{56}}     & 3   & \textcolor{red}{\ding{56}}   & vary    & No  & No \\
          & Unreal~\cite{zhang2021unrealperson} & 3,000  & 120,000 & 34     & \textcolor{red}{\ding{56}}     & \textcolor{red}{\ding{56}}     & 4   & \textcolor{red}{\ding{56}}   & low  & Many  & No \\
          & RandPerson~\cite{wang2020surpassing} & 8,000  & 1,801,816 & 19    & \textcolor{red}{\ding{56}}     & \textcolor{red}{\ding{56}}     & 11   & \textcolor{red}{\ding{56}}   & low    & No  & No \\
    \cline{2-12}      & \textbf{\textit{FineGPR} (Ours)} & \textbf{1,150}  & \textbf{2,028,600} & \textbf{36}    & \textbf{7}     & \textbf{7}     & \textbf{9}   &\textcolor[rgb]{0.00,0.50,0.00}{\ding{52}}   & \textbf{high}    & \textbf{Many}  & \textbf{Addressed} \\
    \Xhline{0.8pt}
    \end{tabular}}%
  \caption{Comparison of some real-world and synthetic person re-ID datasets. In particular, ``\#Wea." , ``\#Illum." , ``\#Scenes"  and ``ID-level" indicate whether dataset has human-annotated labels in terms of weather, illumination, background and ID-level attributes, respectively.}
  \label{tab1}%
\end{table*}%

\section{Related Works}
\subsection{Person re-ID Methods}
In the field of person re-ID, early works~\cite{zhao2014learning,liao2015person} either concentrate on hand-crafted feature or low-level semantic feature. Unfortunately, these methods always fail to produce competitive results because of their limited discriminative learning ability. Recently, benefited from the advances of deep neural networks, person re-ID performance in supervised learning has been significantly boosted to a new level~\cite{wang2016joint,chen2017beyond,li2018harmonious}, which learned robust feature extraction and reliable metric learning in an end-to-end manner. Typically, person re-ID model can be trained with the identification loss~\cite{xiao2017joint}, contrastive loss~\cite{varior2016gated,varior2016siamese} and triplet loss~\cite{hermans2017defense}. Recently, a strong baseline~\cite{luo2019bag} for re-ID is employed to extract the discriminative feature, which has been proved to have great potential to learn a robust and discriminative model in person re-ID models.
Besides, several literatures~\cite{deng2018image,xiang2020unsupervised} focus on the image-level to allow different domains to have similar feature distributions, or adopt an adversarial domain adaptation approach to mitigate the distribution shift~\cite{ganin2015unsupervised,torralba2011unbiased}, which has attracted considerable attention from various fields in re-ID community.

\subsection{Person re-ID Datasets}
Being the foundation of more sophisticated re-ID techniques, the pursuit of better datasets never stops in the area of person re-ID. Early attempts could be traced back to VIPeR~\cite{gray2007evaluating}, ETHZ~\cite{schwartz2009learning} and RAiD~\cite{das2014consistent}. More challenging datasets are proposed subsequently, including Market-1501~\cite{zheng2015scalable}, CUHK03~\cite{li2014deepreid}, MSMT17~\cite{wei2018person}, \textit{etc.} However, labelling such a large-scale real-world dataset is labor-intensive and time-consuming, sometimes there even exists security and privacy problems.
Besides, all of these datasets only have limited attribute distribution and are lack of diversity. As the performance gain is gradually saturated on the above datasets, newly large-scale datasets are needed urgently to further boost re-ID performance.
Recently, leveraging synthetic data is an effective idea to alleviate the reliance on large-scale real-world datasets.
This strategy has been applied in various computer vision tasks, \textit{e.g.}, object detection~\cite{pepik2012teaching}, crowd counting~\cite{wang2019learning} and semantic segmentation~\cite{chen2019learning}.
In the person re-ID community,
many re-ID methods~\cite{barbosa2018looking,bak2018domain,sun2019dissecting,wang2020surpassing} have proposed to take advantage of game engine to construct large-scale synthetic re-ID datasets, which can be used to pre-train or fine-tune CNN network.
For example, Barbosa \textit{et al.}~\cite{barbosa2018looking} propose a synthetic dataset SOMAset created by photo-realistic human body generation software to enrich the diversity. Recently, Wang \textit{et al.}~\cite{wang2020surpassing} collect a virtual dataset RandPerson with 3D characters containing 1,801,816 synthetic images of 8,000 identities. However, these datasets are either in a small scale or lack of diversity, few of them provide rich attribute annotations, which cannot satisfy the need of attribute learning in person re-ID task.
So new fine-grained annotated datasets are urgently needed.

\section{The \textit{FineGPR} Dataset}
In this section, we describe \textit{FineGPR}, a new dataset with high-quality annotations and multiple attribute distributions to re-ID community. Below we at first review the process of constructing and annotation collection,  then present an analysis over
dataset statistics. We show some sample images from the proposed \textit{FineGPR} dataset in Fig.~\ref{fig2}.

\begin{figure*}[!t]
\centerline{\includegraphics[width=0.9\linewidth]{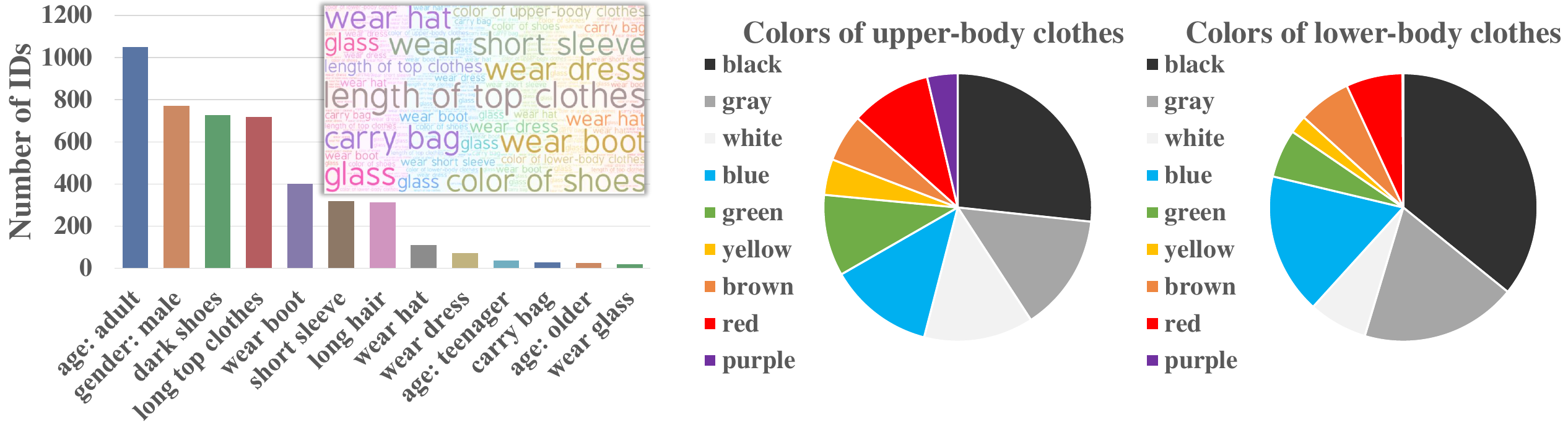}}
\caption{The distributions of attributes at the identity level on \textit{FineGPR}. The left figure shows the numbers of IDs and category cloud for each attribute.
The middle and right pies illustrate the distribution of the colors of upper-body and low-body clothes respectively.}
\label{fig12}
\end{figure*}

\subsection{Dataset Collection}
Our \textit{FineGPR} dataset is collected from a popular game engine called the Grand Theft Auto V (GTA5). Practically, we create a synthetic controllable world containing 2,028,600 synthesized person images of 1,150 identities. Images in this dataset generally contain different attributes in a large scope, \textit{e.g.}, \textit{\textbf{Viewpoint}}, \textit{\textbf{Weather}}, \textit{\textbf{Illumination}}, \textit{\textbf{Background}} and \textit{\textbf{ID-level annotations}}, also including many hard samples with occlusion. It is worth noting that, all images are simultaneously captured by 36 non-overlapping cameras with a high resolution and image quality.
In the process of image generation, each person walks along a schedule route, and cameras are set up and fixed at the chosen locations. As a controllable system, it can satisfy various data requirements in a fine-grained fashion.

\subsection{Properties of \textbf{\textit{FineGPR}} dataset}
The goal of our \textit{FineGPR} dataset is to introduce a new challenging benchmark with high-quality annotations and multiple attribute distribution to re-ID community. To the best of our knowledge, this is the first large-scale person dataset with over 4 environment-level attributes and 13 ID-level attribute annotations.

\textbf{Identities.} According to the Table~\ref{tab1}, \textit{FineGPR} contains 1,150 hand-crafted identities including females and males, with resolution of $200\times480$.
To ensure diversity, we cropped human region with different angles. As shown in Fig.~\ref{fig2} (\textit{second row}), different person has different body shape, clothing, hairstyle, and the motion can be randomly set as walking, running, standing and so on.
Particularly, the clothes of these characters include jeans, pants, shorts, skirts, T-shirts, dress shirts, \textit{etc.}, and some of these identities have a backpack, shoulder bag, and wear glasses or hat. In total, we manually annotate the \textit{FineGPR} with 13 different pedestrian attributes at the identity level (\textit{e.g.}, wearing dress or not), the distribution diagram is demonstrated in Fig.~\ref{fig12}.

\textbf{Viewpoint.}
We construct the image exemplars under specified viewpoints. Those images are randomly sampled during normal walking, running, \textit{etc.} Formally, a person image is sampled every $10^{\circ}$ from $0^{\circ}$ to $350^{\circ}$. (36 different types of viewpoints in total). There are 49 images for each viewpoint of an identity in the entire \textit{FineGPR}, so each person has 1,764 ($49\times36$) images in total.

\textbf{Weather.}
Currently, the proposed \textit{FineGPR} has 7 different weather conditions, including Sunny, Clouds, Overcast, Foggy, Neutral, Blizzard and Snowlight. It is worth mentioning that the number of instances in each weather condition is the same, but not a natural heavy-tail distribution, which makes it adaptable to various real-world scenarios.

\textbf{Illumination.}
Illumination is another critical factor that contributes to the success of generalizable re-ID, which consists of 7 different types of illumination, \textit{e.g.} midnight (time period during 23:00--4:00 in 24 hours a day.), dawn (4:00--6:00), forenoon (6:00--11:00), noon (11:00--13:00), afternoon (13:00--18:00), dusk(18:00--20:00) and night (20:00--23:00).  Parameters like time setting can be modified manually for each illumination type. By editing the values of these terms, various kinds of illumination environments can be created.

\textbf{Background.}
GTA5 has a very large environment map, including thousands of realistic urban areas and wild scenes. From now, 9 different scenes are selected to represent real-world scenarios with annotations, \textit{e.g.} street, mall, school, park and mountain, \textit{etc.}, which are distributed evenly across all identities. The different scenes are shown in Fig.~\ref{fig2} (\textit{first row}). More additional details related to our \textit{FineGPR} can be found in the Supplementary Material.

\textbf{Ethical Considerations.}
People-centric datasets pose some challenges of data privacy~\cite{fabbri2021motsynth} and intersectional accuracy disparities~\cite{buolamwini2018gender}. To address such concerns, our dataset were created with
careful attention to ethical
questions, which we encountered
throughout our work. Access to our dataset will be provided
for research purposes only and with restrictions on redistribution.  Furthermore,
we are very cautious of annotation procedure of \textit{FineGPR} dataset to avoid the social and ethical implications.
As for re-ID system, governments and officials must establish strict regulations to control the usage of this technology since it mainly relies on (not all) surveillance data. Motivated by this,
we do not consider the dataset for developing non-research systems unless further professional processing or augmentation.
\begin{figure*}
\centerline{\includegraphics[width=0.9\linewidth]{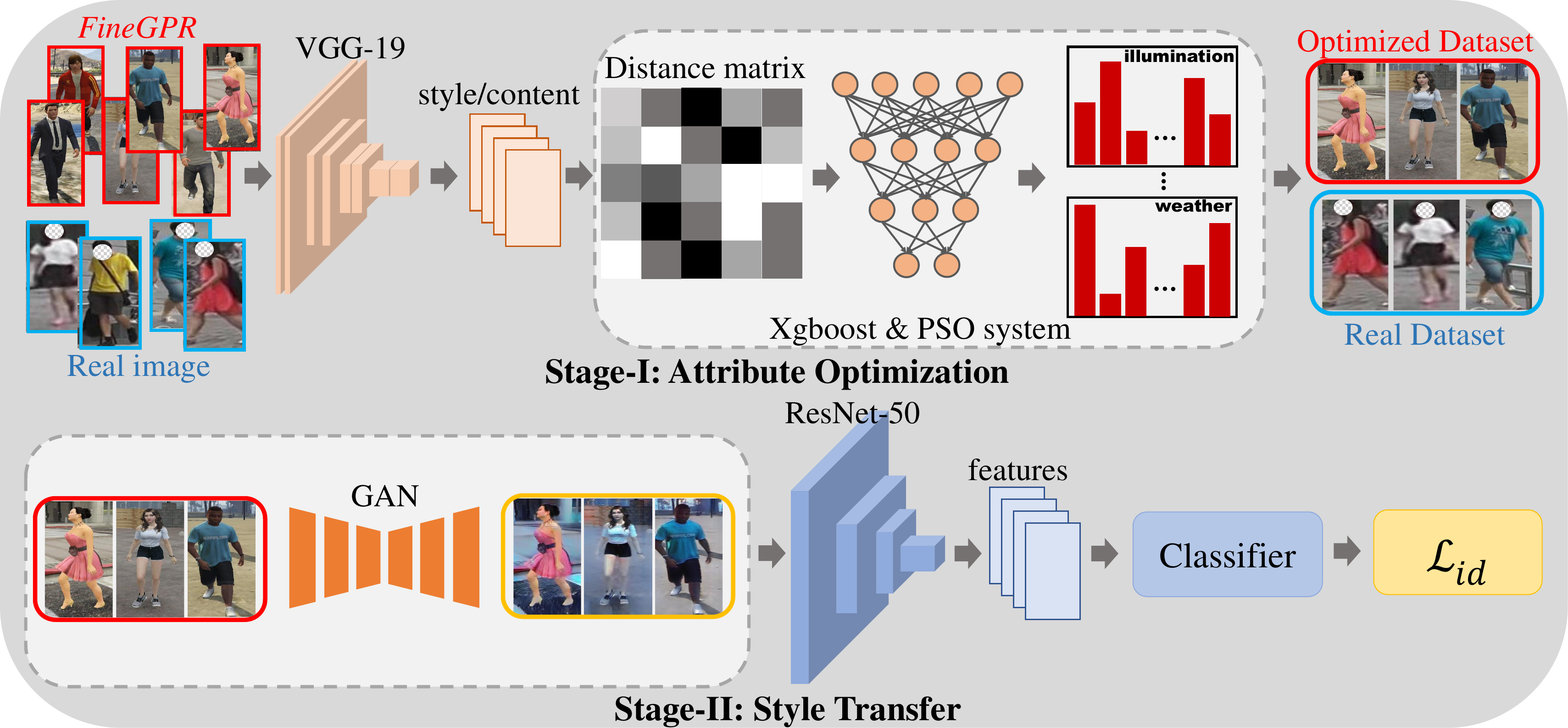}}
\caption{The two-stage pipeline \textbf{AOST} to learn attribute distribution of target domain. Firstly, we learn attribute distribution of real domain on the basis of XGBoost \& PSO learning system. Secondly, we perform style transfer to enhance the reality of optimal dataset. Finally, the transferred data are adopted for downstream re-ID task.}
\label{fig5}
\end{figure*}

\begin{figure}[!t]
\centerline{\includegraphics[width=1.0\linewidth]{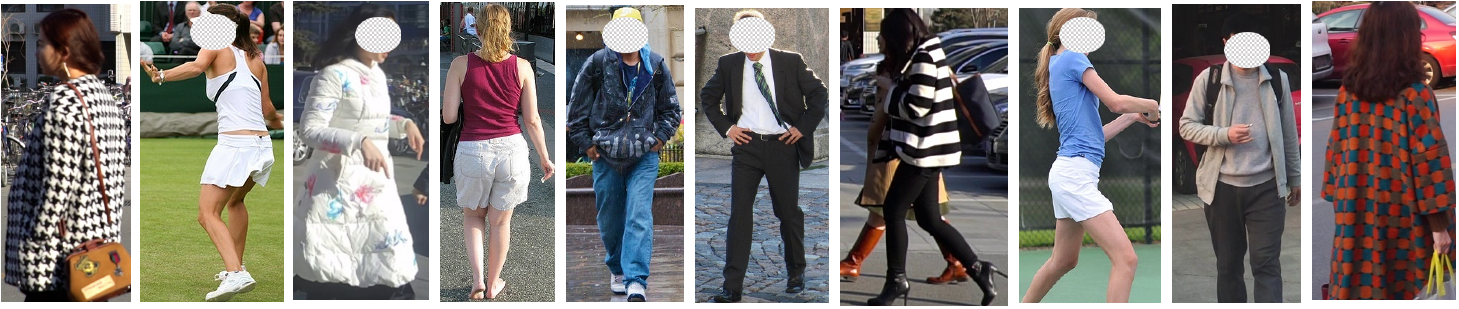}}
\caption{Some visual examples of collected \textbf{MSCO} dataset.}
\label{fig11}
\end{figure}

\section{Methodology of Proposed AOST}
In this section, we design an effective training strategy AOST to directly select samples on the basis of synthetic \textit{FineGPR} for
initializing the re-ID backbone. And the overall framework is illustrated in Fig.~\ref{fig5}, which includes two stages: Attribute Optimization and Style Transfer.

\textbf{Attribute Optimization.}
Intuitively, since \textit{FineGPR} is a large-attribute-range dataset, using entire \textit{FineGPR} for training is time-consuming and low-efficient. To further exploit the potential of \textit{FineGPR} and promote the training efficiency, we introduce a novel strategy to learn some representative attributes with prior target knowledge. Following the procedure in~\cite{gatys2016image}, we adopt a widely used backbone VGG-19~\cite{simonyan2014very} pre-trained on ImageNet~\cite{deng2009imagenet} to obtain the style distance $\mathcal{D}_{\text {style}}$ and content distance $\mathcal{D}_{\text {content}}$ respectively, which are formulated as:
\begin{equation}
\label{eq_content}
\mathcal{D}_{\text {content }}=\frac{1}{2} \sum_{i, j}\left(F_{i j}^{l}-P_{i j}^{l}\right)^{2}
\end{equation}
\begin{equation}\label{eq_style}
\mathcal{D}_{\text {style}}=\sum_{l=0}^{L} w_{l} \frac{1}{4 N_{l}^{2} M_{l}^{2}} \sum_{i, j}\left(G_{i j}^{l}-A_{i j}^{l}\right)^{2}
\end{equation}
where $F_{i j}^{l}$ and $P_{i j}^{l}$ denote representations extracted by the  $i^{th}$ filter at position $j$ in layer $l$ of VGG-19.  $w_{l}$ is a hyper-parameter which controls the importance of each layer to the style . $N_{l}$ represents the number of filters and $M_{l}$ is size of the feature map.
 $G_{i j}$ and $A_{i j}$ denote the Gram Matrix of real and synthetic images in layer $l$.
 Then total distance for attribute metric is represented as
\begin{equation}\label{eq3}
\mathcal{D}_{\text {total}} = \alpha * \mathcal{D}_{\text {style}} + \beta * \mathcal{D}_{\text {content}}
\end{equation}
where $\alpha$ and $\beta$ are two hyper-parameters which control the relative importance of style and content distance respectively. As depicted in Algorithm~\ref{alg1}, a tree boosting system named XGBoost~\cite{chen2016xgboost} model $\theta_{0}$ is trained with \textit{FineGPR} and $\mathcal{D}_{\text {total}}$. Based on the upgraded Xgboost model $\theta^{*}$,
we continuously adopt a wildly used Particle Swarm Optimization (PSO)~\cite{kennedy1995particle} method to search some optimized attributes. Typically, it selects a similar style or content distribution with respect to a real target dataset. The optimization framework can be shown in Fig.~\ref{fig5} (\textbf{Stage I}). To our knowledge this is the first demonstration to perform attribute optimization with large-scale synthetic dataset on person re-ID task. In essence, compared with existing methods for attribute optimization, such as reinforcement learning~\cite{ruiz2018learning} and attribute descent~\cite{yao2020simulating}, our method investigates and learns these attribute distributions only with few parameters to optimize, which makes it more flexible and adaptable.

\renewcommand{\algorithmicrequire}{ \textbf{Input:}} 
\renewcommand{\algorithmicensure}{ \textbf{Output:}} 
\floatname{algorithm}{\textbf{Algorithm}}
\begin{algorithm}[t]
\caption{The Proposed AOST Method}
\small
\label{alg1}
\begin{algorithmic}
 \Require
Labeled synthetic data $L$;
Unlabeled real target data $U$;\\
Initialized VGG model $\phi$;
Xgboost model $\theta_{0}$;\\
Two hyper-parameters $\alpha$ and $\beta$; Iteration rounds \textit{n};
 \Ensure
Best re-ID model $f\left(\boldsymbol{w}, x_{i}\right)$
\label{ code:fram:extract }
\end{algorithmic}
\begin{algorithmic}[1]
\State Initialize: \textit{m} = 1, \textit{iter} = 1;
\State \textcolor[rgb]{0.11,0.21,0.65}{$\rhd$ \textit{Attribute Optimization ***}}
\label{code1}
\State Extract $\mathcal{D}_{\text {style}} \& \mathcal{D}_{\text {content}}$ beween $L$ and $U$ with model $\phi$;

\label{code2}
\State $\mathcal{D}_{\text {total}} \leftarrow \alpha * \mathcal{D}_{\text {style}} + \beta * \mathcal{D}_{\text {content}}$;

\While{$m \leq\|U\| $} \do

\label{code3}
\State Optimized model $\theta^{*}$ $\leftarrow$ train $\theta_{0}$ with $L$ and $\mathcal{D}_{\text {total}}$;

\label{code4}
\State Optimized attributes $\mathcal{V^{*}}$ $\leftarrow$ update $\theta^{*}$ with PSO;

\label{code5}
\State Update the sample size: $m \leftarrow m + 1$;

\EndWhile

\label{code6}
\State Generate a new dataset $L^{*}$ according to $\mathcal{V^{*}}$;

\State \textcolor[rgb]{0.11,0.21,0.65}{$\rhd$ \textit{Style Transfer ***}}
\label{code7}
\State Performing style transfer with GAN on $L^{*}$;

\label{code8}
\If {iter $\leq$ n}
    \State Initializing re-ID model with $L^{*}$ by softmax loss ;
    \State \textit{iter} $\leftarrow$ \textit{iter} + 1;
\EndIf
\end{algorithmic}
\end{algorithm}

\begin{table*}[htbp]
  \centering
  \footnotesize
  \setlength{\tabcolsep}{2.2mm}{
    \begin{tabular}{l|c|c|ccc|ccc|ccc}
    \Xhline{0.8pt}
    \multicolumn{3}{c|}{Testing set $\rightarrow$} & \multicolumn{3}{c|}{Market-1501} & \multicolumn{3}{c|}{MSMT17} & \multicolumn{3}{c}{CUHK03} \bigstrut\\
    \hline\hline
    Training set $\downarrow$  & Reference  & Synthetic data & Rank-1 & Rank-5 & mAP   & Rank-1 & Rank-5 & mAP   & Rank-1 & Rank-5 & mAP \bigstrut\\
    \hline
    Market-1501~\cite{zheng2015scalable}  & ICCV 15  & $\times$     & \textit{\underline{92.7}}    & \textit{\underline{97.9}}    & \textit{\underline{81.4}}    & 6.0  & 11.2  & 1.9  & 5.3   & 12.4  & 6.2 \\
    MSMT17~\cite{wei2018person}  & CVPR 18  & $\times$     & 50.2  & 67.7  & 25.7  & \textit{\underline{75.7}}    & \textit{\underline{86.9}}    & \textit{\underline{51.5}}    & 9.9   & 20.4  & 10.7  \\
    CUHK03~\cite{li2014deepreid}  & CVPR 14  & $\times$     & 36.6  & 53.9  & 16.6  & 4.6  & 9.1  & 1.3   & \textit{\underline{43.6}}    & \textit{\underline{62.9}}    & \textit{\underline{41.5}} \\
    SOMAset$^{*}$~\cite{barbosa2018looking}  & CVIU 18  & \checkmark      & 4.5   & --    & 1.3   & 1.4   & --    & 0.3   & 0.4   & --    & 0.4  \\
    SyRI$^{*}$~\cite{bak2018domain}  & ECCV 18   & \checkmark      & 29.0  & --    & 10.8  & 16.4  & --    & 4.4   & 4.1   & --    & 3.5  \\
    Unreal$^\ddagger$~\cite{zhang2021unrealperson}  & CVPR 21   & \checkmark      & 37.4  & 55.2    & 15.9  & 3.9  & 7.4    & 1.3  & 4.3   & 10.0    & 4.7 \\
    PersonX$^{*}$~\cite{sun2019dissecting}  & CVPR 19  & \checkmark      & 44.0  & --    & 20.4  & 11.7  & --    & 3.6  & 7.4   & --    & 6.2  \\
    RandPerson$^{*}$~\cite{wang2020surpassing}  & MM 20   & \checkmark      & \textcolor{blue}{\textbf{55.6}}  & --    & \textcolor{blue}{\textbf{28.8}}  & \textcolor{red}{\textbf{20.1}}  & --    & \textcolor{red}{\textbf{6.3}}  & \textcolor{blue}{\textbf{13.4}}  & --    & \textcolor{blue}{\textbf{10.8}}  \\
    \hline
    \textit{FineGPR}  & Ours  & \checkmark     & 50.5  & 67.7  & 24.6  & 12.5  & 18.5  & 3.9  & 8.7   & 18.2  & 8.4  \\
    \rowcolor{mygray}
    \textit{FineGPR}$^\dagger$   & Ours   & \checkmark     & \textcolor{red}{\textbf{56.3}}  & \textcolor{red}{\textbf{70.4}}  & \textcolor{red}{\textbf{29.2}}  & \textcolor{blue}{\textbf{19.7}}  & \textcolor{red}{\textbf{27.4}}  & \textcolor{blue}{\textbf{6.1}}  & \textcolor{red}{\textbf{14.2}}  & \textcolor{red}{\textbf{20.6}}  & \textcolor{red}{\textbf{11.2}} \\
    \Xhline{0.8pt}
    \end{tabular}%
  \caption{Performance comparison with existing Real and Synthetic datasets on Market-1501, MSMT17 and CUHK03, respectively. \textcolor{red}{\textbf{Red}} indicates the best and \textcolor{blue}{\textbf{Blue}} the second best. $^{*}$ means results are reported by RandPerson~\cite{wang2020surpassing}. $^\ddagger$ represents results reproduced with Unreal\_v2.1 on our baseline. \textit{\underline{Underline}} denotes supervised learning. $^\dagger$ means performing selecting with our AOST method.}
  \label{tab2}}%
\end{table*}%
\textbf{Style Transfer.}
In the above attribute optimization stage, there exists serious domain gap or distribution shift between synthetic and real-world scenario.  Generative Adversarial Networks (GAN)~\cite{goodfellow2014generative} which have demonstrated impressive results on image-to-image translation seem to be a natural solution to this problem. However, existing methods are both inefficient and ineffective in practical application. Their inefficiency results from the fact that a new generator needs to be retrained when given a new real-world scenario. Meanwhile, these methods mainly employ low-resolution images to train a generator, and they are incapable of fully exploiting the potential of GAN, which is likely to limit the quality of generated images.
To provide a remedy to this dilemma, we build a high-resolution dataset MSCO and crawled over 20K images with a size of nearly $200\times480$, which mainly from COCO~\cite{lin2014microsoft} dataset and few from other real-world person datasets. Different locations are also considered to cover a large diversity. We believe that a unified dataset with high-resolution can provide more useful and discriminative information during translation. Some visual examples of collected
MSCO dataset are illustrated in Fig.~\ref{fig11}.
By doing so, we only need to train one generator and translate the synthetic images into photo-realistic style at testing phase.
The details can be seen in Fig.~\ref{fig5} (\textbf{Stage II}).
To verify the priority of MSCO, we adopt several state-of-the-art methods for style-level domain adaptation, \textit{e.g.}, CycleGAN~\cite{zhu2017unpaired}, PTGAN~\cite{wei2018person} and SPGAN~\cite{deng2018image}.

\section{Experiments}
\subsection{Datasets and Evaluation}
\textbf{Market-1501}~\cite{zheng2015scalable} contains 32,668 labeled images of 1,501 identities captured from campus in Tsinghua University.  Each identity is captured by at most 6 cameras. The training set contains 12,936 images from 751 identities and the test set contains 19,732 images from 750 identities.

\textbf{MSMT17}~\cite{wei2018person} has 126,441 labeled images belonging to 4,101 identities and contains 32,621 training images from 1,041 identities. For the testing set, 11,659 bounding boxes are used as query images and other 82,161
bounding boxes are used as gallery images.

\textbf{CUHK03}~\cite{li2014deepreid} contains 14,097 images of 1,467 identities.
Following the CUHK03-NP protocol~\cite{zhong2017re}, it is divided into 7,365 images of 767 identities as the training set, and the remaining 6,732 images of 700 identities as the testing set.
We adopt mean Average Precision (mAP) and Cumulative Matching Characteristics (CMC) at rank-1 and rank-5 for evaluation on re-ID task.

\subsection{Experiment Settings}
We mainly use the newly-built \textit{FineGPR} to conduct the experiments. For attribute optimization, we empirically set
$w_{l}=0.2$ in Eq.~\ref{eq_style}, and $\alpha=0.9$, $\beta=1$ in Eq.~\ref{eq3}.
It is worth mentioning that our re-ID baseline system is built only with commonly used softmax cross-entropy loss~\cite{zhang2018generalized} on vanilla ResNet-50~\cite{he2016deep} with no bells and whistles. Following the practice in~\cite{luo2019bag}, person images are resized to 256 $\times$ 128, then a random horizontal flipping with 0.5 probability is used for data augmentation. The batch size of training samples is set as 128.
Adam method~\cite{kingma2014adam} is adopted for optimization. The initial learning rate is set to 3.5$\times$10$^{-4}$ for the backbone network. Then, these learning rates are decayed to 3.5$\times$10$^{-5}$ and 3.5$\times$10$^{-6}$ at 40th epoch and 70th epoch respectively, and the training stops after 120 epochs.

\subsection{Comparison with the State-of-the-arts}
\label{sec4.2}
To evaluate the superiority of our synthetic dataset, we perform training on \textit{FineGPR} and testing on each individual real dataset. The evaluation results are reported in Table~\ref{tab2}. Surprisingly, when initializing with whole \textit{FineGPR} dataset, we can achieve a rank-1 accuracy of \textbf{50.5\%}, \textbf{12.5\%} and \textbf{8.7\%} when tested on Market-1501, MSMT17 and CUHK03 respectively.
Although there is a slight inferiority of performance when compared with RandPerson~\cite{wang2020surpassing}, our \textit{FineGPR} selected by AOST with fine-grained attributes can lead a significant improvement by \textbf{+0.7\%} and \textbf{+0.8\%} in rank-1 accuracy on Market-1501 and CUHK03 dataset respectively. When compared with real-world datasets, \textit{FineGPR} also outperforms these benchmarks by an impressively large margin in terms of rank-1 accuracy, leading \textbf{+0.3\%} and \textbf{+13.9\%} improvement on Market-1501 compared with MSMT17 and CUHK03 separately. However, initializing with whole \textit{FineGPR} dataset is time-consuming and low-efficient, this motivates the investigation of data selecting techniques that can potentially address this problem.

\begin{table*}[htbp]
  \centering
  \footnotesize
  \setlength{\tabcolsep}{1.5mm}{
    \begin{tabular}{l|c|c|ccc|ccc|ccc}
    \Xhline{0.8pt}
    \multicolumn{3}{c|}{Testing set $\rightarrow$} & \multicolumn{3}{c|}{Market-1501} & \multicolumn{3}{c|}{MSMT17} & \multicolumn{3}{c}{CUHK03} \bigstrut\\
    \hline\hline
    Training set $\downarrow$      & Bboxes    & Time (GPU-days) $\downarrow$    & Rank-1 & Rank-5 & mAP  & Rank-1 & Rank-5 & mAP   & Rank-1 & Rank-5 & mAP \bigstrut\\
    \hline
    \textit{FineGPR}    & 2,028,600    & 20     & 50.5  & 67.7  & 24.6  & 12.5  & 18.5  & 3.9  & 8.7   & 18.2  & 8.4  \\
    \textit{FineGPR}+R   & 124,200    & 1.3     & 39.9  & 57.1  & 18.3  & 4.6  & 9.7  & 1.4  & 5.9   & 15.4  & 5.4  \\
    \textit{FineGPR}+AO (\textit{w/o} transfer)     & 124,200    & 1.3     & 45.5  & 63.2  & 23.8  & 9.1  & 14.7  & 3.1  & 8.5   & 16.9  & 8.3  \\
    \hline\hline
    PersonX+R+ST   & 124,200    & 1.3     & 28.7  & 45.9  & 11.8  & 7.1  & 13.4  & 2.1  & 3.1   & 7.4  & 3.1  \\
    Unreal+R+ST   & 124,200    & 1.3     & 42.8  & 59.3  & 18.4  & 11.3  & 20.5  & 3.5  & 5.4   & 12.6  & 5.1  \\
    RandPerson+R+ST    & 124,200    & 1.3     & 51.4  & 68.3  & 25.0  & 15.8  & 26.7  & 5.0  & 8.4   & 18.1  & 7.5  \\
    \hline
    \rowcolor{mygray}
    \textit{FineGPR}+AOST (Ours)   & \textbf{124,200}    & \textbf{1.3}     & \textbf{56.3}  & \textbf{70.4}  & \textbf{29.2}  & \textbf{19.7}  & \textbf{27.4}  & \textbf{6.1}  & \textbf{14.2}  & \textbf{20.6}  & \textbf{11.2} \\
    \Xhline{0.8pt}
    \end{tabular}}%
  \caption{Controlled experiments by different regulations of our proposed AOST method on Market-1501, MSMT17 and CUHK03, respectively. ``R" indicates random sampling, ``AO" represents Attribute Optimization. ``ST" means Style Transfer.}
  \label{tab3}%
\end{table*}%

\begin{figure*}[t]
\begin{minipage}[t]{0.249\linewidth}
\centering
\includegraphics[width=1.88in]{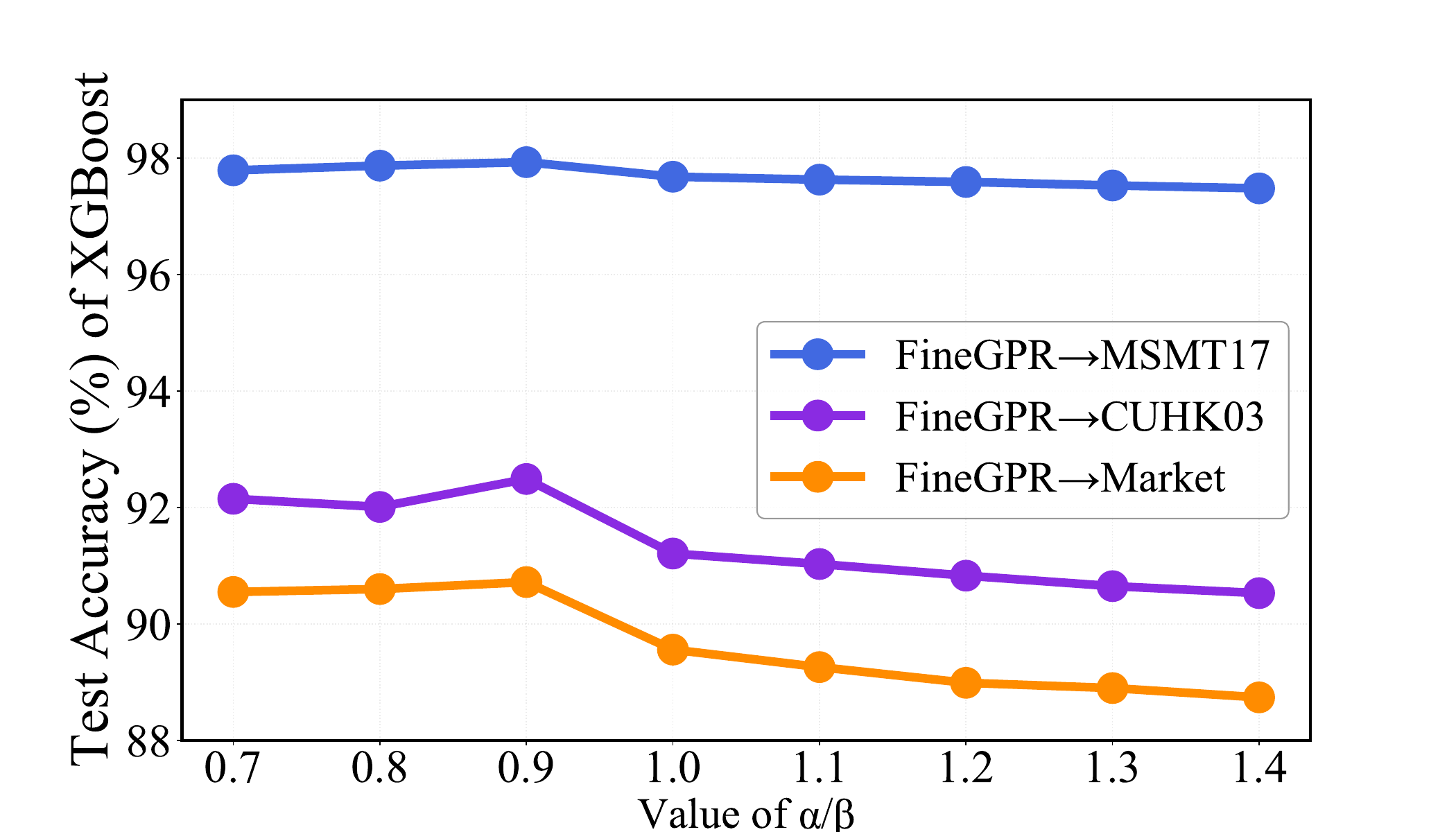}
\centerline{(a) Sensitivity to $\alpha / \beta$}
\label{figa1}
\end{minipage}%
\begin{minipage}[t]{0.249\linewidth}
\centering
\includegraphics[width=1.88in]{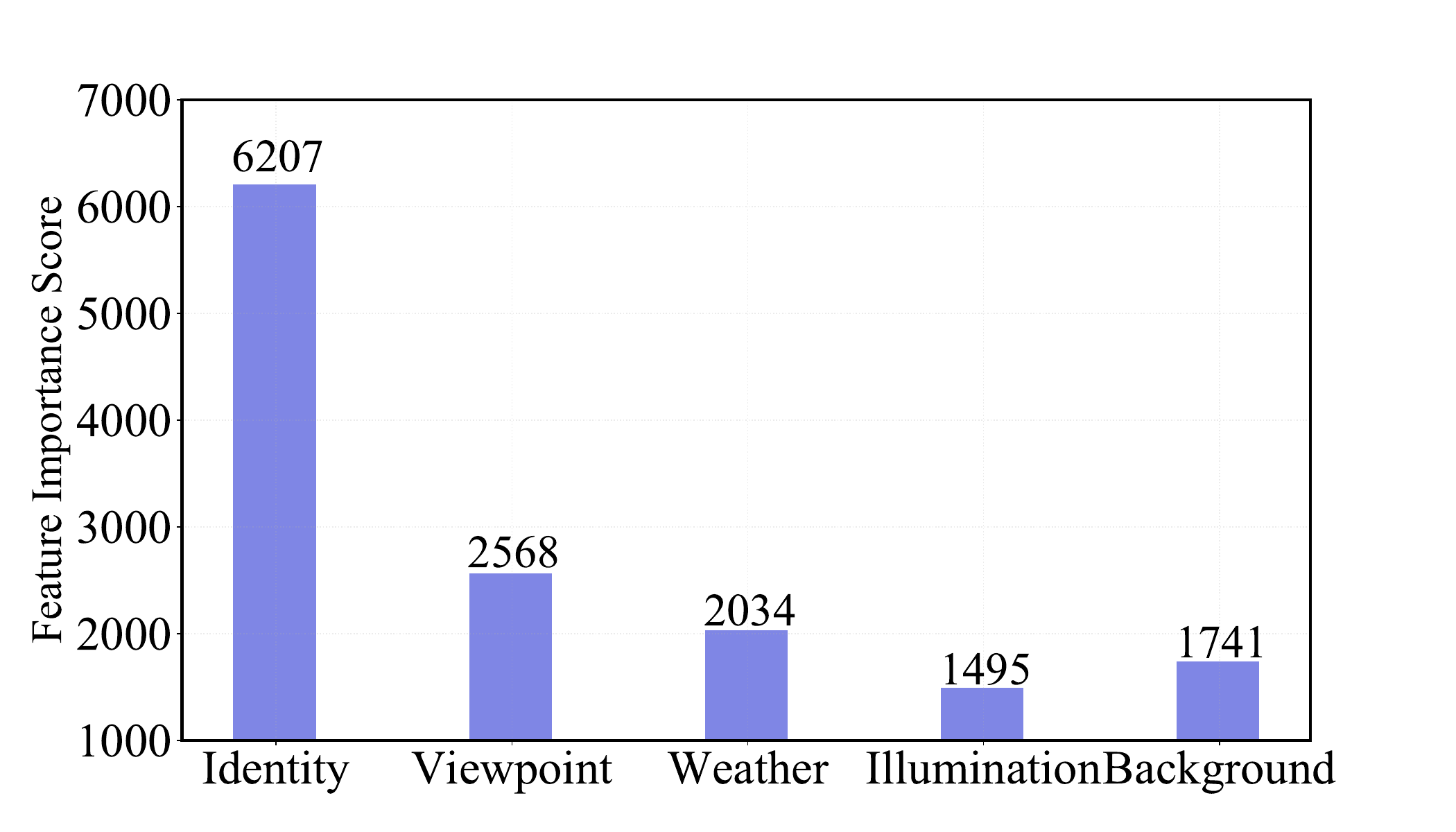}
\centerline{(b) \textit{FineGPR}$\rightarrow$Market}
\label{figa1}
\end{minipage}%
\begin{minipage}[t]{0.249\linewidth}
\centering
\includegraphics[width=1.88in]{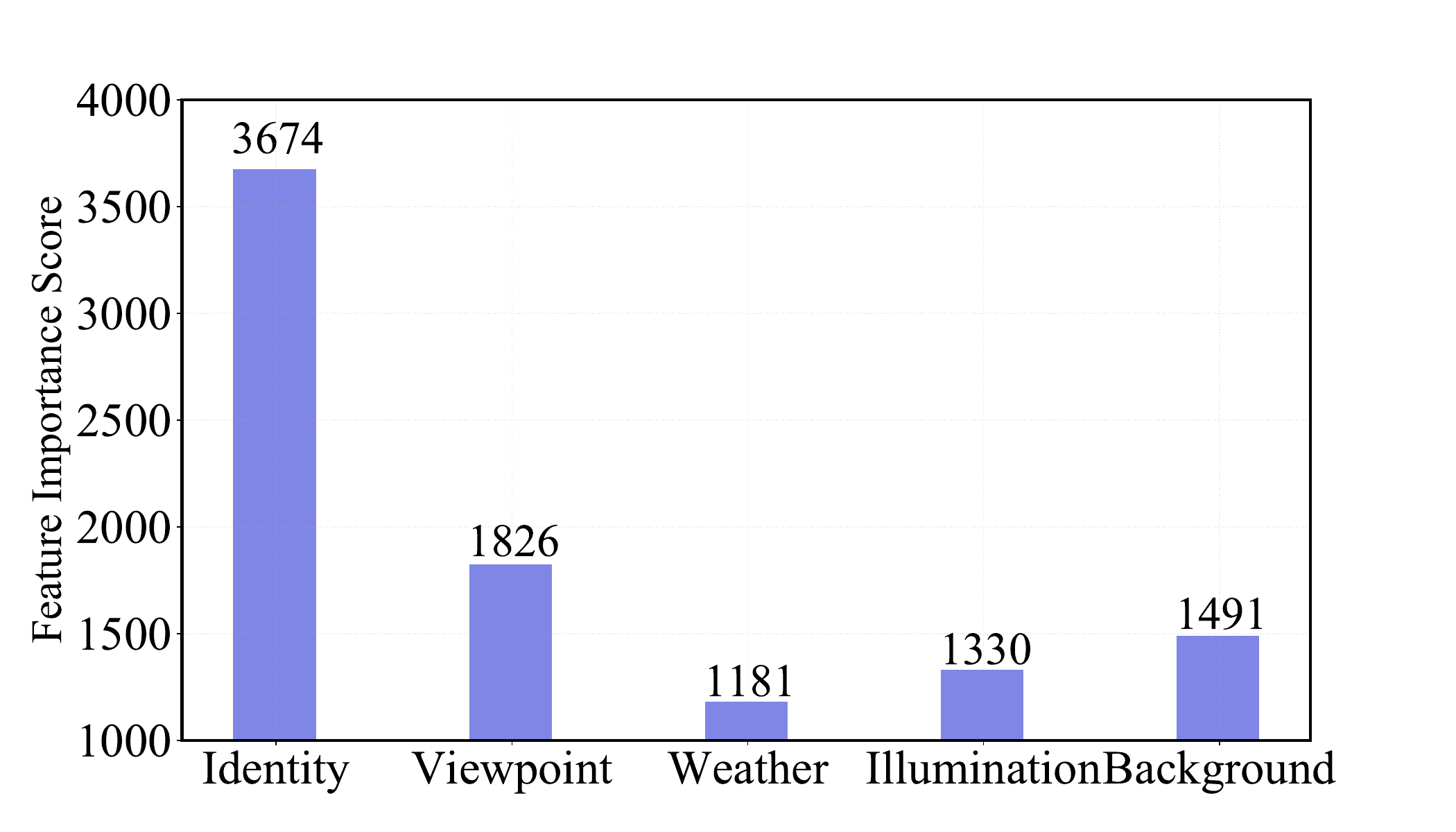}
\centerline{(c) \textit{FineGPR}$\rightarrow$MSMT17}
\label{figa2}
\end{minipage}
\begin{minipage}[t]{0.249\linewidth}
\centering
\includegraphics[width=1.88in]{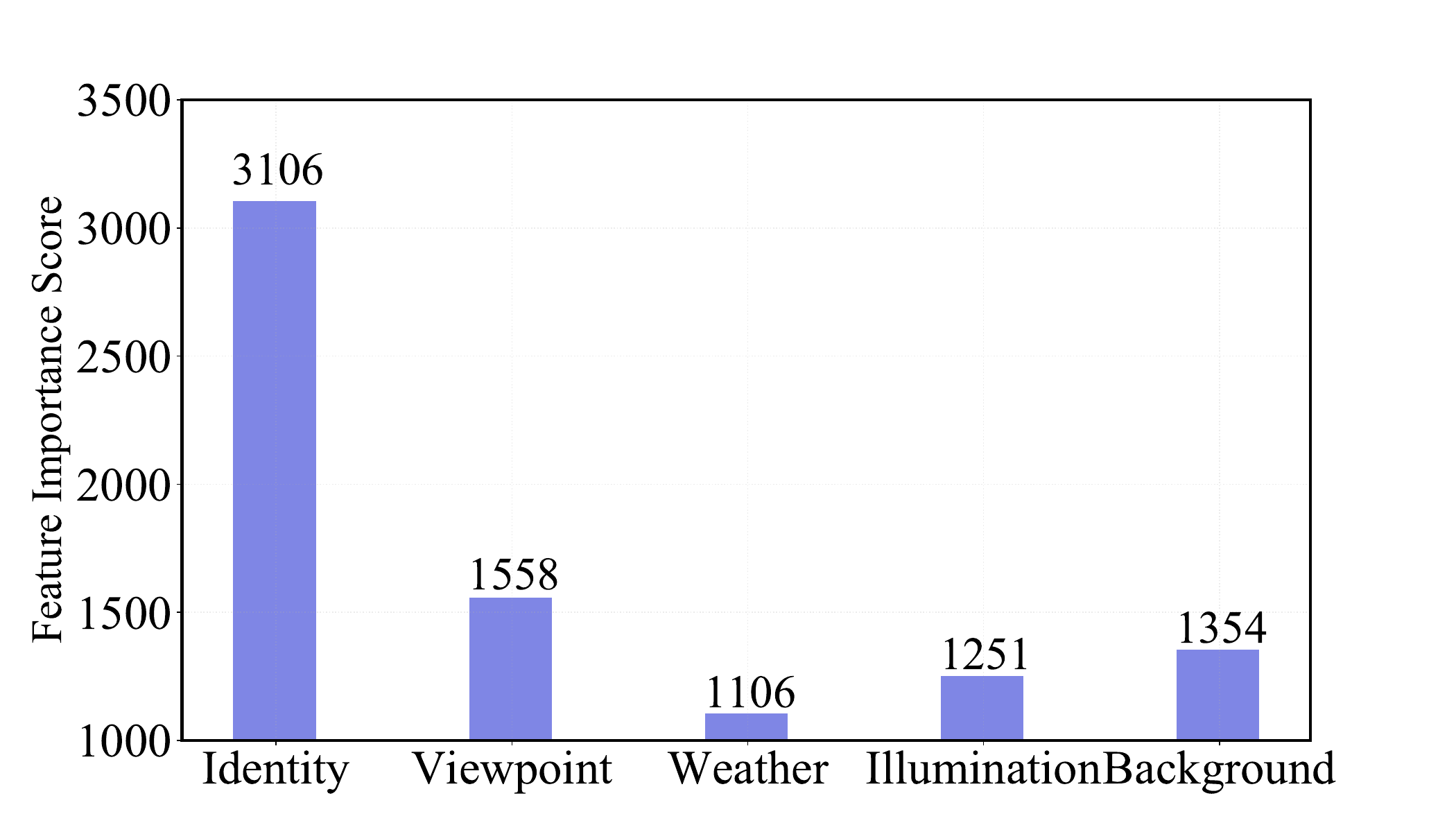}
\centerline{(d) \textit{FineGPR}$\rightarrow$CUHK03}
\label{figa3}
\end{minipage}%
\caption{(a) Sensitivity of AOST to key parameter $\alpha / \beta$ in Eq.~\ref{eq3}. (b-d) Feature importance of XGBoost with Feature Importance Score (higher is more important) on Market-1501, MSMT17 and CUHK03 respectively. Please zoom in for the best view.}
\label{fig13}
\end{figure*}

\subsection{Ablation Study}
\textbf{Important Parameters.}
In Eq.~\ref{eq3}, both $\alpha$ and $\beta$  controls the relative importance of the style and content distance respectively between real and synthetic samples. Since $\mathcal{D}_{\text {total}}$ is a linear combination between $\mathcal{D}_{\text {style}}$ and $\mathcal{D}_{\text {content}}$, we can smoothly regulate the emphasis or adjust the trade-off between the content or the style.
As depicted in Fig.~\ref{fig13} (a), we observe that when $\alpha / \beta$ is small, the performances is not optimal because the style representation is way too limited to a very small portion, and thus our AOST could only mine the discriminative information in terms of content representation of the re-ID data. The $\alpha / \beta$ should also not be set too large, otherwise the performances  drops dramatically since the model mine too many samples in style representation. Specially, $\alpha / \beta=0.9$ yields the best accuracy.

\textbf{Evaluation of Attribute Importance.}
Based on the end-to-end AOST system, we evaluate the impacts of different attributes in a fine-grained manner by the
XGBoost in terms of \textit{gain}~\cite{chen2016xgboost} (feature importance score). As illustrated in Fig.~\ref{fig13} (b-d), it can be easily observed that \textbf{Identity} accounts for the largest proportion no matter which real dataset is employed for testing, followed by \textbf{Viewpoint} attribute. Typically, this conclusion is in accordance with our prior knowledge on generalizable re-ID problem, that is, \textit{using more IDs as training samples is always beneficial to the re-ID system, and viewpoint also plays a key role in recognizing the clothes appearance.}
More details about the importance related to ID-level attributes is provided in the Supplementary. We hope these analysis about attribute importance will provide useful insights
for dataset building and future practical usage to the community.

\textbf{Effectiveness of Attribute Optimization.}
We proceed study on dependency by testing whether the Attribute Optimization (AO) matters. According to Table~\ref{tab3}, our attribute optimization strategy \textit{FineGPR}+AO (\textit{w/o} transfer) can lead a
significant improvement in rank-1 of \textbf{+5.6\%}, \textbf{+4.5\%} and \textbf{+2.6\%} on Market-1501, MSMT17 and CUHK03 respectively when compared with random sampling (\textit{FineGPR}+R). We suspect this is due to samples selected by attribute optimization strategy are much closer to real target domain, and the learned attribute distribution has a higher quality, which have a direct impact on downstream re-ID task.
Meanwhile, fast training is our second main advantage since the scale of training set can be largely decreased by attribute optimization,  \textit{e.g.}, it costs nearly 20 GPU-days\footnote{All timings use one Nvidia Tesla P100 GPU on a server equipped with a Intel Xeon E5-2690 V4 CPU.} when pre-training on entire \textit{FineGPR} with 2,028,600 images. Fortunately, training time will be considerably reduced by \textbf{15$\times$} (\textcolor[rgb]{0.20,0.40,0.80}{\textbf{20}} vs. \textcolor{red}{\textbf{1.3}} GPU-days) by our proposed AOST without performance degradation, which leads a more efficient deployment to real-world scenarios. Surprisingly, even with fewer samples for training, our approach still yields its
competitiveness when compared with existing datasets, \textit{e.g.} \textcolor{red}{\textbf{56.3\%}} vs. \textcolor{blue}{\textbf{55.6\%}} in rank-1 on Market in Table~\ref{tab2}, proving the proverbial \textit{\textbf{less-is-more}} principle.

To go even further, we also adopt AOST on synthetic datasets to prove the priority of \textit{FineGPR}. Unfortunately, due to lack of fine-grained attribute annotations, these datasets (\textit{e.g.} PersonX, Unreal and RandPerson) cannot
satisfy the need for AOST in re-ID. We instead randomly sample 124,200 images from
these datasets individually, and then perform style transfer.
As illustrated in Table~\ref{tab3}, it can be easily observed that \textit{FineGPR}+AOST can perform significantly greater than PersonX+R+ST, Unreal+R+ST and RandPerson+R+ST separately, which successfully proves the applicability of our proposed dataset and approach.

\textbf{Effectiveness of Style Transfer.}
Synthetic data engine can generate images and annotations at lower labor costs. However, there exists obvious domain gap between synthetic and real-world scenario, which hinders the further improvement in performance on downstream task.
Note that for training efficiency, we instead consider an easier but practical strategy, that is, employing off-the-shelf style transfer model to generate photo-realistic images for further effective training. As shown in Table~\ref{tab3}, \textit{w/o} style transfer by AOST, the rank-1 accuracy drops sharply from 56.3\% to 45.5\% and the mAP drops from 29.2\% to 23.8\%.
\textit{This confirms that mitigating domain gap between synthetic and real dataset is a crucial ingredient to make the performance to an excellent level.} For simplicity, the SPGAN is used as the style transfer model in the following experiments.

\begin{figure}[!t]
\centerline{\includegraphics[width=0.93\linewidth]{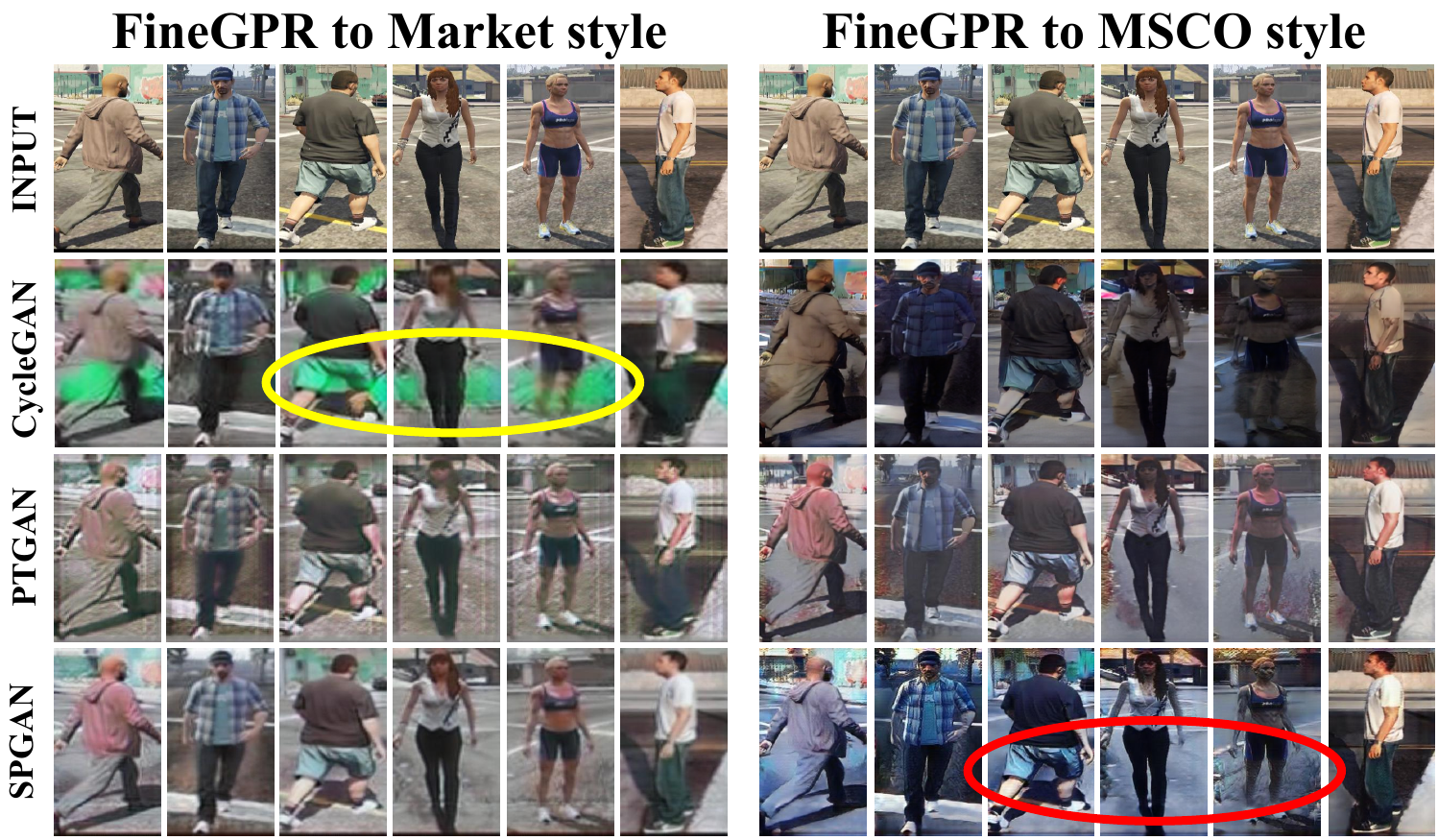}}
\caption{Qualitative comparisons of different GAN methods when trained on Market and our MSCO, respectively.}
\label{fig6}
\end{figure}

\subsection{Qualitative and Quantitative Results}
\textbf{Qualitative Evaluations.}  Fig.~\ref{fig6} presents a visual comparison of different transfer methods trained on low-resolution  Market  and high-resolution MSCO respectively. It can be easily noticed that GANs create artifacts and coarse results (indicated by \textcolor[rgb]{0.80,0.80,0.00}{yellow box}) when trained on low-resolution images, which still remains problematic.
In comparison, our method with MSCO can successfully address the artifacts and produce most visually pleasant results (indicated by \textcolor{red}{red box}) in an even better fashion, which is implicitly beneficial to the downstream re-ID mission.

\begin{figure}[!t]
\centerline{\includegraphics[width=0.95\linewidth]{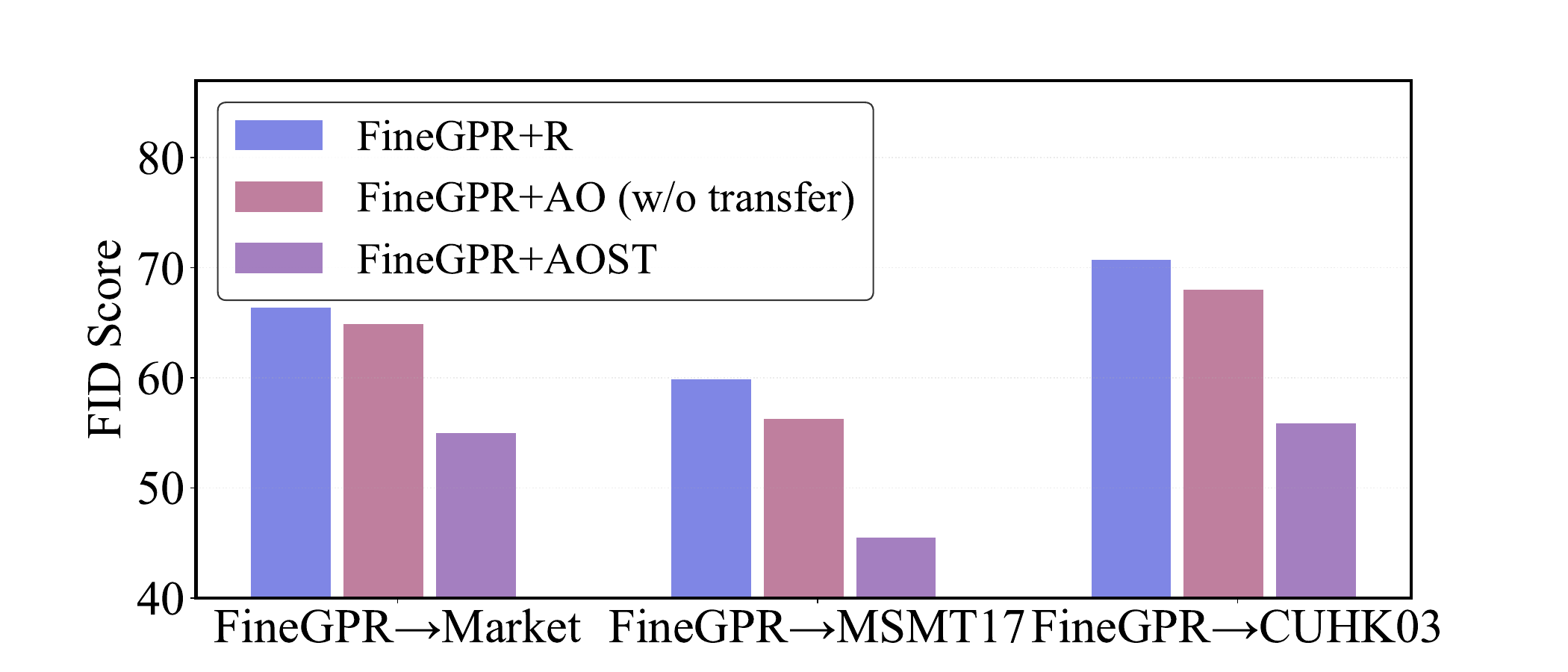}}
\caption{Comparison of FID (lower is better) to evaluate the effectiveness of different regulation terms of AOST.}
\label{fig8}
\end{figure}

\begin{figure}[!t]
\centerline{\includegraphics[width=0.95\linewidth]{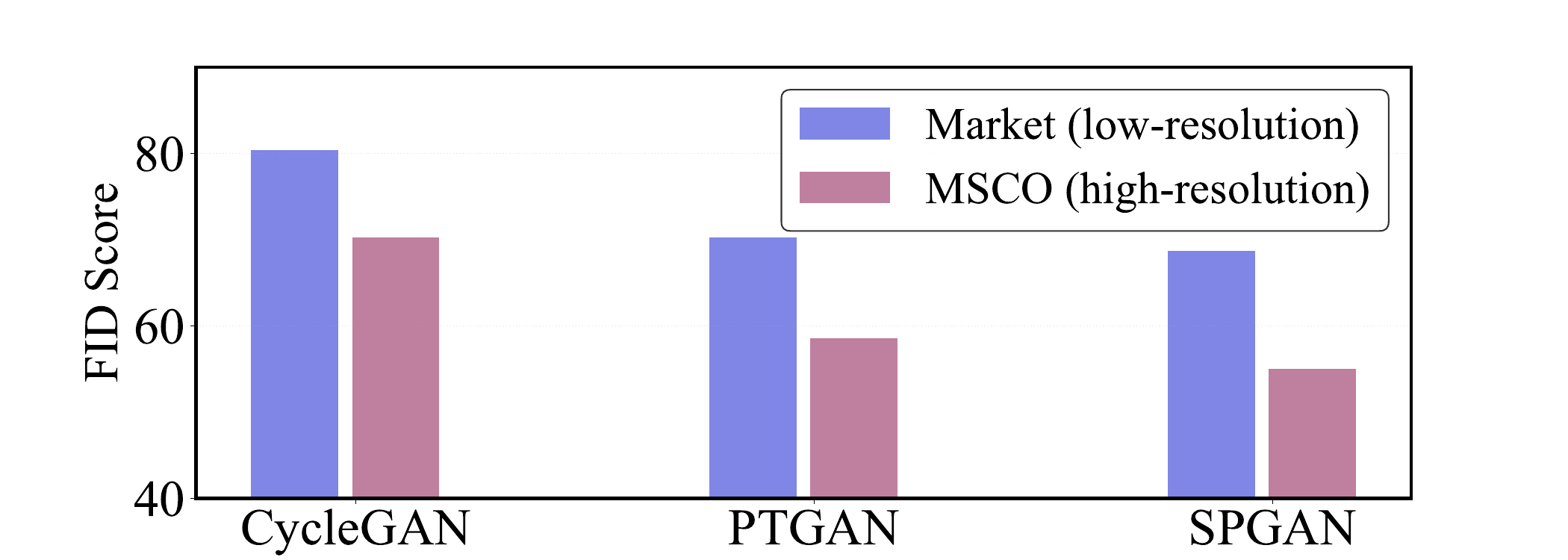}}
\caption{Comparison of FID (lower is better) to evaluate the realism of generated images by CycleGAN, PTGAN and SPGAN when trained
on samples with different resolution.}
\label{fig10}
\end{figure}

\textbf{Quantitative Evaluations.}
Our qualitative observations above are confirmed by the quantitative evaluations. To be more specific, we adopt
Fréchet Inception Distance (FID)~\cite{heusel2017gans} to measure the distribution difference between synthetic and real photos. Generally, FID measures how close the distribution of generated images is to the real. As shown in Fig.~\ref{fig8}, by adding new regulation terms, \textit{e.g.} attribute optimization or style transfer, the FID score gradually decreases no matter which dataset is employed for evaluation, suggesting the learned attribute distributions are more and more similar to the real images. Even prior to this point, according to Fig.~\ref{fig10}, training a generator with low-resolution images always produce low-quality images (indicated by higher FID score) no matter which style transfer model is employed. Still, SPGAN and PTGAN rank the best, while SPGAN shows a slight quantitative advantage. In all, the introduction of high-resolution MSCO dataset can always improve the adaptability to style changes and mitigate previously mentioned domain gap effectively, even in much more complex scenarios.


\section{Conclusion}
In this work, we take the first step to construct the largest person dataset \textit{FineGPR} with fine-grained attribute labels and high-quality annotations. On top of \textit{FineGPR}, we introduce an attribute analysis methodology called AOST to learn important attribute distribution, which enjoys the benefits of small-scale dataset for more efficient training. Continuously, style transfer is adopted to further mitigate domain gap between synthetic and real photos. With this, we proved, for the first time, that a model trained on limited synthetic data can yield a competitive performance in generalizable re-ID task. Extensive experiments also demonstrate the superiority of \textit{FineGPR} and effectiveness of AOST. We hope our dataset and method will shed light into potential tasks for the community to move forward.


{\small
\bibliographystyle{ieee_fullname}
\bibliography{egbib}

\begin{thebibliography}{10}\itemsep=-1pt

\bibitem{bak2018domain}
Slawomir Bak, Peter Carr, and Jean-Francois Lalonde.
\newblock Domain adaptation through synthesis for unsupervised person
  re-identification.
\newblock In {\em Proceedings of the European Conference on Computer Vision
  (ECCV)}, pages 189--205, 2018.

\bibitem{barbosa2018looking}
Igor~Barros Barbosa, Marco Cristani, Barbara Caputo, Aleksander Rognhaugen, and
  Theoharis Theoharis.
\newblock Looking beyond appearances: Synthetic training data for deep cnns in
  re-identification.
\newblock {\em Computer Vision and Image Understanding}, 167:50--62, 2018.

\bibitem{buolamwini2018gender}
Joy Buolamwini and Timnit Gebru.
\newblock Gender shades: Intersectional accuracy disparities in commercial
  gender classification.
\newblock In {\em Conference on fairness, accountability and transparency},
  pages 77--91. PMLR, 2018.

\bibitem{chen2016xgboost}
Tianqi Chen and Carlos Guestrin.
\newblock Xgboost: A scalable tree boosting system.
\newblock In {\em Proceedings of the 22nd acm sigkdd international conference
  on knowledge discovery and data mining}, pages 785--794, 2016.

\bibitem{chen2017beyond}
Weihua Chen, Xiaotang Chen, Jianguo Zhang, and Kaiqi Huang.
\newblock Beyond triplet loss: a deep quadruplet network for person
  re-identification.
\newblock In {\em Proceedings of the IEEE conference on computer vision and
  pattern recognition}, pages 403--412, 2017.

\bibitem{chen2019learning}
Yuhua Chen, Wen Li, Xiaoran Chen, and Luc~Van Gool.
\newblock Learning semantic segmentation from synthetic data: A geometrically
  guided input-output adaptation approach.
\newblock In {\em Proceedings of the IEEE/CVF Conference on Computer Vision and
  Pattern Recognition}, pages 1841--1850, 2019.

\bibitem{das2014consistent}
Abir Das, Anirban Chakraborty, and Amit~K Roy-Chowdhury.
\newblock Consistent re-identification in a camera network.
\newblock In {\em European conference on computer vision}, pages 330--345.
  Springer, 2014.

\bibitem{deng2009imagenet}
Jia Deng, Wei Dong, Richard Socher, Li-Jia Li, Kai Li, and Li Fei-Fei.
\newblock Imagenet: A large-scale hierarchical image database.
\newblock In {\em 2009 IEEE conference on computer vision and pattern
  recognition}, pages 248--255. Ieee, 2009.

\bibitem{deng2018image}
Weijian Deng, Liang Zheng, Qixiang Ye, Guoliang Kang, Yi Yang, and Jianbin
  Jiao.
\newblock Image-image domain adaptation with preserved self-similarity and
  domain-dissimilarity for person re-identification.
\newblock In {\em Proceedings of the IEEE conference on computer vision and
  pattern recognition}, pages 994--1003, 2018.

\bibitem{fabbri2021motsynth}
Matteo Fabbri, Guillem Bras{\'o}, Gianluca Maugeri, Orcun Cetintas, Riccardo
  Gasparini, Aljosa Osep, Simone Calderara, Laura Leal-Taixe, and Rita
  Cucchiara.
\newblock Motsynth: How can synthetic data help pedestrian detection and
  tracking?
\newblock In {\em Proceedings of the IEEE/CVF International Conference on
  Computer Vision}, pages 10849--10859, 2021.

\bibitem{ganin2015unsupervised}
Yaroslav Ganin and Victor Lempitsky.
\newblock Unsupervised domain adaptation by backpropagation.
\newblock In {\em International conference on machine learning}, pages
  1180--1189. PMLR, 2015.

\bibitem{gatys2016image}
Leon~A Gatys, Alexander~S Ecker, and Matthias Bethge.
\newblock Image style transfer using convolutional neural networks.
\newblock In {\em Proceedings of the IEEE conference on computer vision and
  pattern recognition}, pages 2414--2423, 2016.

\bibitem{goddard2017eu}
Michelle Goddard.
\newblock The eu general data protection regulation (gdpr): European regulation
  that has a global impact.
\newblock {\em International Journal of Market Research}, 59(6):703--705, 2017.

\bibitem{goodfellow2014generative}
Ian Goodfellow, Jean Pouget-Abadie, Mehdi Mirza, Bing Xu, David Warde-Farley,
  Sherjil Ozair, Aaron Courville, and Yoshua Bengio.
\newblock Generative adversarial nets.
\newblock {\em Advances in neural information processing systems}, 27, 2014.

\bibitem{gray2007evaluating}
Douglas Gray, Shane Brennan, and Hai Tao.
\newblock Evaluating appearance models for recognition, reacquisition, and
  tracking.
\newblock In {\em Proc. IEEE international workshop on performance evaluation
  for tracking and surveillance (PETS)}, volume~3, pages 1--7. Citeseer, 2007.

\bibitem{he2016deep}
Kaiming He, Xiangyu Zhang, Shaoqing Ren, and Jian Sun.
\newblock Deep residual learning for image recognition.
\newblock In {\em Proceedings of the IEEE conference on computer vision and
  pattern recognition}, pages 770--778, 2016.

\bibitem{hermans2017defense}
Alexander Hermans, Lucas Beyer, and Bastian Leibe.
\newblock In defense of the triplet loss for person re-identification.
\newblock {\em arXiv preprint arXiv:1703.07737}, 2017.

\bibitem{heusel2017gans}
Martin Heusel, Hubert Ramsauer, Thomas Unterthiner, Bernhard Nessler, and Sepp
  Hochreiter.
\newblock Gans trained by a two time-scale update rule converge to a local nash
  equilibrium.
\newblock {\em Advances in neural information processing systems}, 30, 2017.

\bibitem{kennedy1995particle}
James Kennedy and Russell Eberhart.
\newblock Particle swarm optimization.
\newblock In {\em Proceedings of ICNN'95-international conference on neural
  networks}, volume~4, pages 1942--1948. IEEE, 1995.

\bibitem{kingma2014adam}
Diederik~P Kingma and Jimmy Ba.
\newblock Adam: A method for stochastic optimization.
\newblock {\em arXiv preprint arXiv:1412.6980}, 2014.

\bibitem{li2014deepreid}
Wei Li, Rui Zhao, Tong Xiao, and Xiaogang Wang.
\newblock Deepreid: Deep filter pairing neural network for person
  re-identification.
\newblock In {\em Proceedings of the IEEE conference on computer vision and
  pattern recognition}, pages 152--159, 2014.

\bibitem{li2018harmonious}
Wei Li, Xiatian Zhu, and Shaogang Gong.
\newblock Harmonious attention network for person re-identification.
\newblock In {\em Proceedings of the IEEE conference on computer vision and
  pattern recognition}, pages 2285--2294, 2018.

\bibitem{liao2015person}
Shengcai Liao, Yang Hu, Xiangyu Zhu, and Stan~Z Li.
\newblock Person re-identification by local maximal occurrence representation
  and metric learning.
\newblock In {\em Proceedings of the IEEE conference on computer vision and
  pattern recognition}, pages 2197--2206, 2015.

\bibitem{lin2014microsoft}
Tsung-Yi Lin, Michael Maire, Serge Belongie, James Hays, Pietro Perona, Deva
  Ramanan, Piotr Doll{\'a}r, and C~Lawrence Zitnick.
\newblock Microsoft coco: Common objects in context.
\newblock In {\em European conference on computer vision}, pages 740--755.
  Springer, 2014.

\bibitem{luo2019bag}
Hao Luo, Youzhi Gu, Xingyu Liao, Shenqi Lai, and Wei Jiang.
\newblock Bag of tricks and a strong baseline for deep person
  re-identification.
\newblock In {\em Proceedings of the IEEE/CVF Conference on Computer Vision and
  Pattern Recognition Workshops}, pages 0--0, 2019.

\bibitem{pan2018two}
Xingang Pan, Ping Luo, Jianping Shi, and Xiaoou Tang.
\newblock Two at once: Enhancing learning and generalization capacities via
  ibn-net.
\newblock In {\em Proceedings of the European Conference on Computer Vision
  (ECCV)}, pages 464--479, 2018.

\bibitem{pepik2012teaching}
Bojan Pepik, Michael Stark, Peter Gehler, and Bernt Schiele.
\newblock Teaching 3d geometry to deformable part models.
\newblock In {\em 2012 IEEE conference on computer vision and pattern
  recognition}, pages 3362--3369. IEEE, 2012.

\bibitem{ristani2016performance}
Ergys Ristani, Francesco Solera, Roger Zou, Rita Cucchiara, and Carlo Tomasi.
\newblock Performance measures and a data set for multi-target, multi-camera
  tracking.
\newblock In {\em European conference on computer vision}, pages 17--35.
  Springer, 2016.

\bibitem{ruiz2018learning}
Nataniel Ruiz, Samuel Schulter, and Manmohan Chandraker.
\newblock Learning to simulate.
\newblock {\em arXiv preprint arXiv:1810.02513}, 2018.

\bibitem{schwartz2009learning}
William~Robson Schwartz and Larry~S Davis.
\newblock Learning discriminative appearance-based models using partial least
  squares.
\newblock In {\em 2009 XXII Brazilian symposium on computer graphics and image
  processing}, pages 322--329. IEEE, 2009.

\bibitem{simonyan2014very}
Karen Simonyan and Andrew Zisserman.
\newblock Very deep convolutional networks for large-scale image recognition.
\newblock {\em arXiv preprint arXiv:1409.1556}, 2014.

\bibitem{sun2019dissecting}
Xiaoxiao Sun and Liang Zheng.
\newblock Dissecting person re-identification from the viewpoint of viewpoint.
\newblock In {\em Proceedings of the IEEE/CVF Conference on Computer Vision and
  Pattern Recognition}, pages 608--617, 2019.

\bibitem{torralba2011unbiased}
Antonio Torralba and Alexei~A Efros.
\newblock Unbiased look at dataset bias.
\newblock In {\em CVPR 2011}, pages 1521--1528. IEEE, 2011.

\bibitem{varior2016gated}
Rahul~Rama Varior, Mrinal Haloi, and Gang Wang.
\newblock Gated siamese convolutional neural network architecture for human
  re-identification.
\newblock In {\em European conference on computer vision}, pages 791--808.
  Springer, 2016.

\bibitem{varior2016siamese}
Rahul~Rama Varior, Bing Shuai, Jiwen Lu, Dong Xu, and Gang Wang.
\newblock A siamese long short-term memory architecture for human
  re-identification.
\newblock In {\em European conference on computer vision}, pages 135--153.
  Springer, 2016.

\bibitem{wang2016joint}
Faqiang Wang, Wangmeng Zuo, Liang Lin, David Zhang, and Lei Zhang.
\newblock Joint learning of single-image and cross-image representations for
  person re-identification.
\newblock In {\em Proceedings of the IEEE Conference on Computer Vision and
  Pattern Recognition}, pages 1288--1296, 2016.

\bibitem{wang2019learning}
Qi Wang, Junyu Gao, Wei Lin, and Yuan Yuan.
\newblock Learning from synthetic data for crowd counting in the wild.
\newblock In {\em Proceedings of the IEEE/CVF Conference on Computer Vision and
  Pattern Recognition}, pages 8198--8207, 2019.

\bibitem{wang2020surpassing}
Yanan Wang, Shengcai Liao, and Ling Shao.
\newblock Surpassing real-world source training data: Random 3d characters for
  generalizable person re-identification.
\newblock In {\em Proceedings of the 28th ACM International Conference on
  Multimedia}, pages 3422--3430, 2020.

\bibitem{wei2018person}
Longhui Wei, Shiliang Zhang, Wen Gao, and Qi Tian.
\newblock Person transfer gan to bridge domain gap for person
  re-identification.
\newblock In {\em Proceedings of the IEEE conference on computer vision and
  pattern recognition}, pages 79--88, 2018.

\bibitem{xiang2020unsupervised}
Suncheng Xiang, Yuzhuo Fu, Guanjie You, and Ting Liu.
\newblock Unsupervised domain adaptation through synthesis for person
  re-identification.
\newblock In {\em 2020 IEEE International Conference on Multimedia and Expo
  (ICME)}, pages 1--6. IEEE, 2020.

\bibitem{xiao2017joint}
Tong Xiao, Shuang Li, Bochao Wang, Liang Lin, and Xiaogang Wang.
\newblock Joint detection and identification feature learning for person
  search.
\newblock In {\em Proceedings of the IEEE Conference on Computer Vision and
  Pattern Recognition}, pages 3415--3424, 2017.

\bibitem{yao2020simulating}
Yue Yao, Liang Zheng, Xiaodong Yang, Milind Naphade, and Tom Gedeon.
\newblock Simulating content consistent vehicle datasets with attribute
  descent.
\newblock In {\em Computer Vision--ECCV 2020: 16th European Conference,
  Glasgow, UK, August 23--28, 2020, Proceedings, Part VI 16}, pages 775--791.
  Springer, 2020.

\bibitem{zhang2021unrealperson}
Tianyu Zhang, Lingxi Xie, Longhui Wei, Zijie Zhuang, Yongfei Zhang, Bo Li, and
  Qi Tian.
\newblock Unrealperson: An adaptive pipeline towards costless person
  re-identification.
\newblock In {\em Proceedings of the IEEE/CVF Conference on Computer Vision and
  Pattern Recognition}, pages 11506--11515, 2021.

\bibitem{zhang2018generalized}
Zhilu Zhang and Mert~R Sabuncu.
\newblock Generalized cross entropy loss for training deep neural networks with
  noisy labels.
\newblock In {\em 32nd Conference on Neural Information Processing Systems
  (NeurIPS)}, 2018.

\bibitem{zhao2014learning}
Rui Zhao, Wanli Ouyang, and Xiaogang Wang.
\newblock Learning mid-level filters for person re-identification.
\newblock In {\em Proceedings of the IEEE conference on computer vision and
  pattern recognition}, pages 144--151, 2014.

\bibitem{zheng2015scalable}
Liang Zheng, Liyue Shen, Lu Tian, Shengjin Wang, Jingdong Wang, and Qi Tian.
\newblock Scalable person re-identification: A benchmark.
\newblock In {\em Proceedings of the IEEE international conference on computer
  vision}, pages 1116--1124, 2015.

\bibitem{zheng2017unlabeled}
Zhedong Zheng, Liang Zheng, and Yi Yang.
\newblock Unlabeled samples generated by gan improve the person
  re-identification baseline in vitro.
\newblock In {\em Proceedings of the IEEE international conference on computer
  vision}, pages 3754--3762, 2017.

\bibitem{zhong2017re}
Zhun Zhong, Liang Zheng, Donglin Cao, and Shaozi Li.
\newblock Re-ranking person re-identification with k-reciprocal encoding.
\newblock In {\em Proceedings of the IEEE conference on computer vision and
  pattern recognition}, pages 1318--1327, 2017.

\bibitem{zhu2017unpaired}
Jun-Yan Zhu, Taesung Park, Phillip Isola, and Alexei~A Efros.
\newblock Unpaired image-to-image translation using cycle-consistent
  adversarial networks.
\newblock In {\em Proceedings of the IEEE international conference on computer
  vision}, pages 2223--2232, 2017.

\end{thebibliography}


\begin{thebibliography}{1}\itemsep=-1pt

\bibitem{fang2017rmpe}
Hao-Shu Fang, Shuqin Xie, Yu-Wing Tai, and Cewu Lu.
\newblock Rmpe: Regional multi-person pose estimation.
\newblock In {\em ICCV}, pages 2334--2343, 2017.

\bibitem{lin2019improving}
Yutian Lin, Liang Zheng, Zhedong Zheng, Yu Wu, Zhilan Hu, Chenggang Yan, and Yi
  Yang.
\newblock Improving person re-identification by attribute and identity
  learning.
\newblock {\em Pattern Recognition}, 95:151--161, 2019.

\bibitem{xiang2021vtbr}
Suncheng Xiang, Zirui Zhang, Mengyuan Guan, Hao Chen, Binjie Yan, Ting Liu, and
  Yuzhuo Fu.
\newblock Vtbr: Semantic-based pretraining for person re-identification.
\newblock {\em arXiv preprint arXiv:2110.05074}, 2021.

\bibitem{zhao2017pyramid}
Hengshuang Zhao, Jianping Shi, Xiaojuan Qi, Xiaogang Wang, and Jiaya Jia.
\newblock Pyramid scene parsing network.
\newblock In {\em CVPR}, pages 2881--2890, 2017.

\end{thebibliography}
}

\end{document}


\title{Supplementary Material for ``Less is More: Learning from Synthetic Data with Fine-grained Attributes for Person Re-Identification"}

\author{Suncheng Xiang$^{1}$, Guanjie You$^{2}$, Mengyuan Guan$^{1}$, Hao Chen$^{1}$, Binjie Yan$^{1}$, Ting Liu$^{1}$, Yuzhuo Fu$^{1}$\\
$^{1}$Shanghai Jiao Tong University, $^{2}$National University of Defense Technology\\
{\tt\small \{xiangsuncheng17, gemini.my, 958577057, yanbinjie, louisa\_liu, yzfu\}@sjtu.edu.cn, ygjssxz@163.com}
}


\maketitle
\begin{abstract}
      This supplementary material accompanies our main manuscript ``Less is More: Learning from Synthetic Data with Fine-grained Attributes  for Person Re-Identification", including more visual results related to our \textit{FineGPR} dataset in terms of viewpoint, weather,
   illumination and background distribution, as well as ID-level attributes,  and more analysis experiment results of our proposed AOST, which related to the \textcolor{red}{Section 3}, \textcolor{red}{Section 4} and \textcolor{red}{Section 5} in the main manuscript respectively.
   In addition, \textit{FineGPR} is not limited to the specific research field of person re-ID, it can also be extended to broader research areas, including other human related task such as segmentation and pose estimation. We hope this dataset could open a door to related research of future tasks.
\end{abstract}

\begin{figure}[!t]
\centerline{\includegraphics[width=0.95\linewidth]{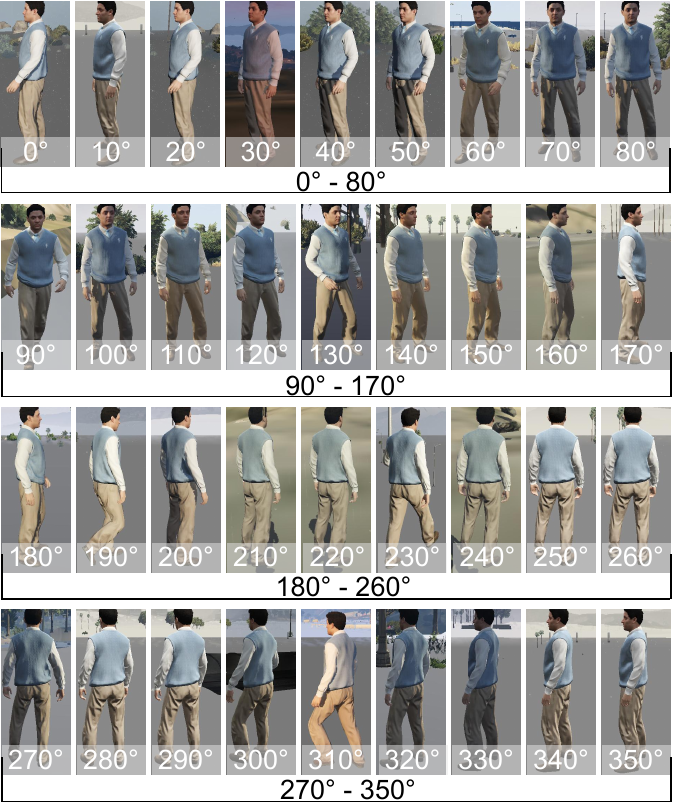}}
\caption{Definition of different viewpoints. Viewpoints of one identity are sampled at an interval of $10^{\circ}$, \textit{e.g.} $0^{\circ}$ $\sim$ $80^{\circ}$ denotes that a person has 9 different angles in total.}
\label{fig2}
\end{figure}

\begin{figure*}[!t]
\centerline{\includegraphics[width=0.95\linewidth]{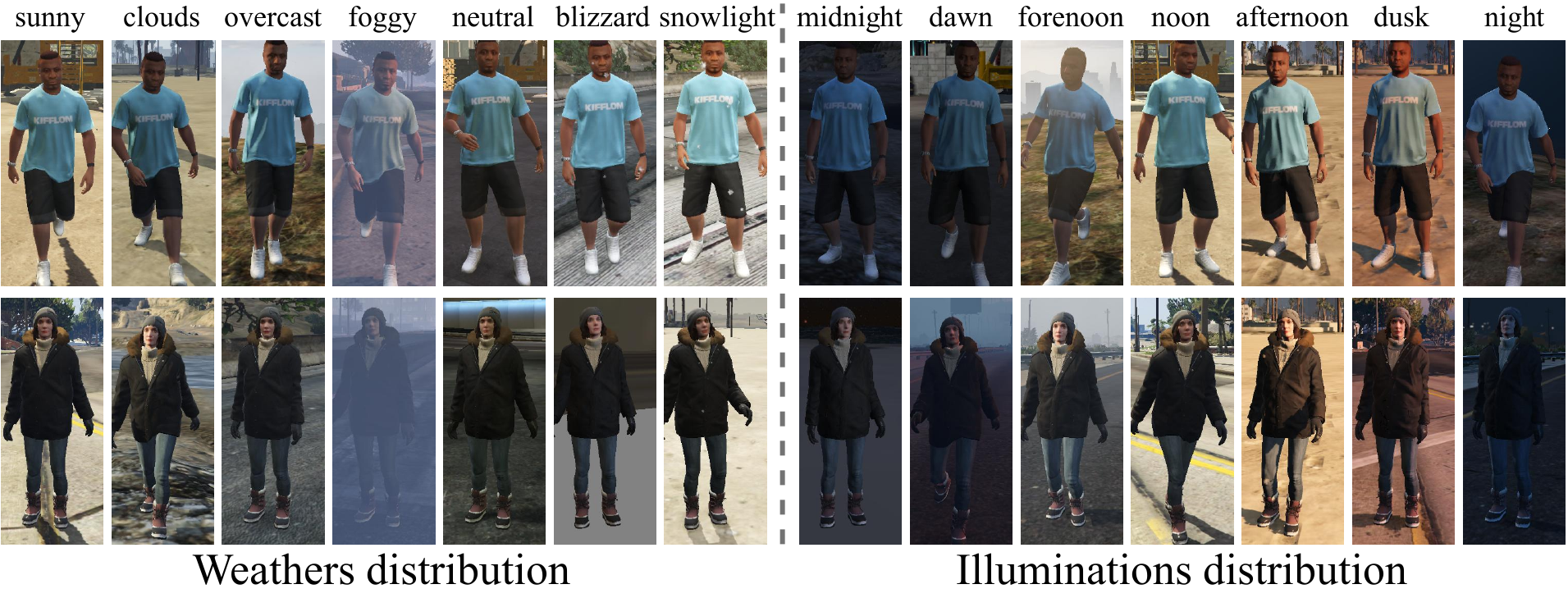}}
\caption{The exemplars of different weather distribution (\textbf{left}) and illumination distribution (\textbf{right}) from the proposed \textit{FineGPR} dataset.}
\label{fig1}
\end{figure*}

\begin{figure}[!t]
\centerline{\includegraphics[width=0.95\linewidth]{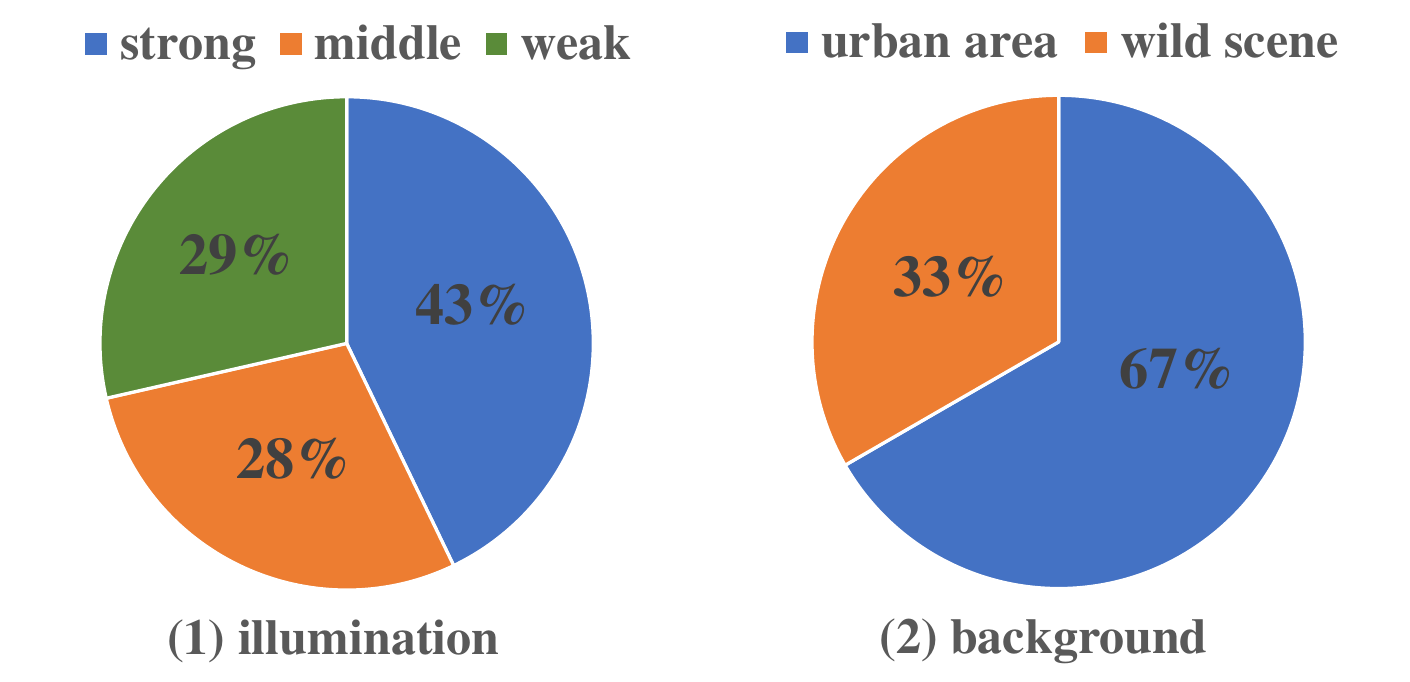}}
\caption{The statistical information of proposed \textit{FineGPR} on illumination (\textbf{strong} light, \textbf{middle} light and \textbf{weak} light) and background (\textbf{urban area} and \textbf{wild scene}) distribution.}
\label{fig14}
\end{figure}

\section{More Demonstrations of \textit{FineGPR} Dataset}
\begin{figure*}[!t]
\centerline{\includegraphics[width=1.0\linewidth]{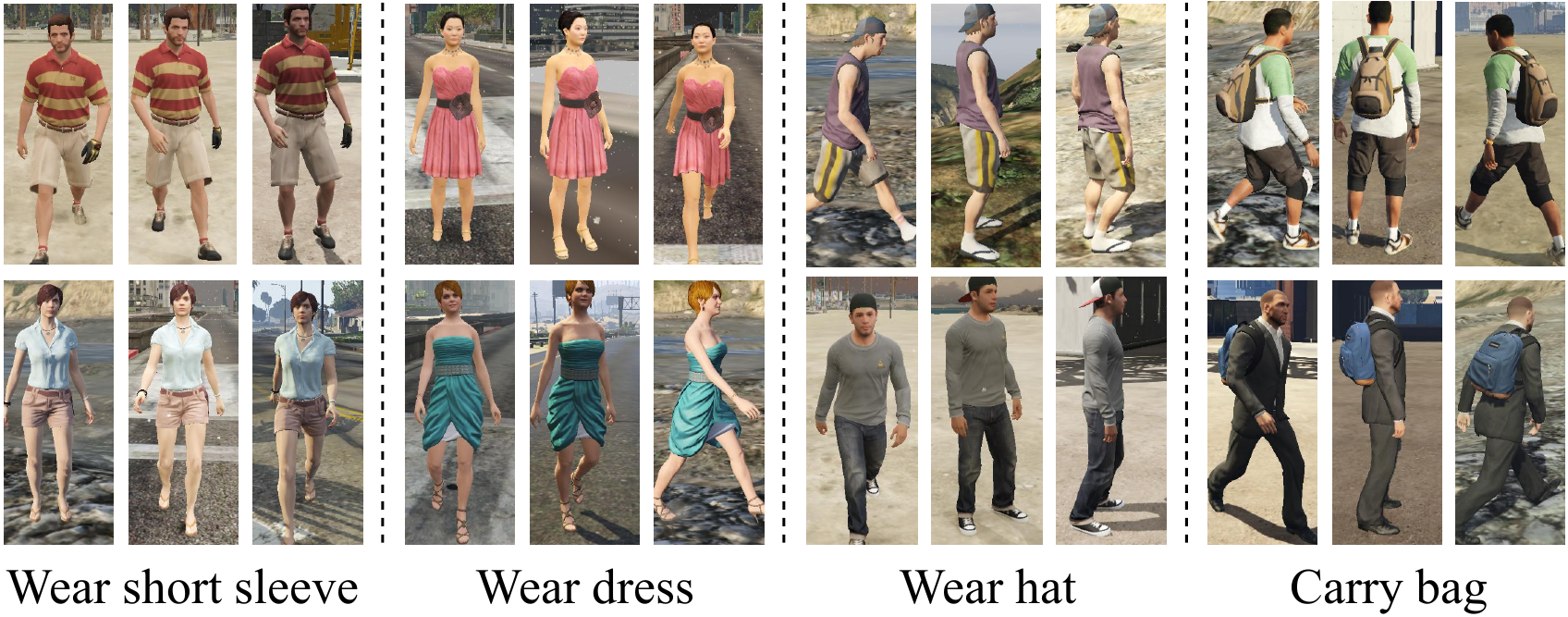}}
\caption{Some visual exemplars with ID-level pedestrian attributes in the proposed \textit{FineGPR} dataset, such as Wear short sleeve , Wear dress, Wear hat, Carry bag, \textit{etc.}.}
\label{fig11}
\end{figure*}

In this section, we will give more visual examples on the distribution of viewpoint, weather, illumination and background attributes of our proposed \textit{FineGPR} dataset in the environment level, as well as multiple pedestrian attributes at the identity level, which will help us better understand the advantages of \textit{FineGPR}. It is worth mentioned that all attributes at the environment level are distributed evenly across all identities in \textit{FineGPR}, which is not heavily biased, but a uniform distribution, \textit{e.g. viewpoint, weather, illumination and background}. So every identity has 36 different viewpoints, 7 different types of weather, 7 different types of illumination and 9 different scenes.\\
\textbf{Viewpoint.} We present the image exemplars under 36 different types of viewpoints. To be more specific, a person image is sampled
every $10^{\circ}$ from $0^{\circ}$ to $350^{\circ}$. A image examples under those different types of viewpoints is demonstrated in Fig.~\ref{fig2}.\\
\textbf{Weather.} For a deeper understanding of our dataset, we present a brief demonstration of  \textit{FineGPR} which consists of 7 different types of weather, \textit{e.g.} \textit{sunny, clouds, overcast, foggy, neutral, blizzard and snowlight}, some visual examples are shown in Fig.~\ref{fig1} (\textbf{left}). These attributes are thoroughly studied in the following experiments.\\
\textbf{Illumination.} As illustrated in Fig.~\ref{fig1} (\textbf{right}), we present some exemplars in different illuminations on top of \textit{FineGPR}, which consists of 7 different types of illumination, \textit{e.g.}, \textit{midnight} (23:00--4:00), \textit{dawn} (4:00--6:00), \textit{forenoon} (6:00--11:00), \textit{noon} (11:00--13:00), \textit{afternoon} (13:00--18:00), \textit{dusk}(18:00--20:00) and \textit{night} (20:00--23:00). Intuitively, we classify these illumination distributions into three categories: \textbf{Weak Light:} midnight, night; \textbf{Middle Light:} dawn, dusk; \textbf{Strong Light:} forenoon, noon, afternoon. All these attributes are carefully designed and provide the new exploration space for fine-grained attribute analysis in re-ID task.\\
\textbf{Locations in GTA5 World}
We have marked the position of each location of our 9 backgrounds in GTA5 world, as shown in Fig.~\ref{fig7}. In particular,
considering the fact that real-world scenarios or existing benchmark datasets in re-ID task are mainly located in the urban scene, so our locations are mainly concentrated in the \textbf{urban area:} street, mall, school, underground parking; few of them are \textbf{wild scene:} park, beach and mountain. The population distribution diagram of our \textit{FineGPR} dataset is shown in Fig.~\ref{fig14}.
Those considerations result in complex backgrounds and scene variations, also make \textit{FineGPR} more appealing and challenging, \textit{etc.} In the future, we will continue to enrich our \textit{FineGPR} datasets by selecting more representative locations and adding more human identities.\\
\textbf{Identity-level Attributes.} Inspired by the work~\cite{lin2019improving}, our synthetic dataset \textit{FineGPR} also provides pedestrian attribute label at the identity level, the category name for all
images in the \textit{FineGPR} dataset is illustrated in Table~\ref{tab5}. The attributes are carefully selected and manually annotated according to the characteristics of the \textit{FineGPR}, which can be used to further boost the performance of generalizable re-ID with attention mechanism. To be more specific, we have labeled 13 different attribute annotations at the identity level: \textit{Gender} (male, female); \textit{hair length} (long, short); \textit{Age} (teenager, adult, older); \textit{Wear glass} (yes, no);
\textit{Wear short sleeve} (yes, no); \textit{length of top clothes} (yes, no); \textit{Wear dress} (yes, no); \textit{Wear boot} (yes, no); \textit{Wear hat} (yes, no);
\textit{Carry bag} (yes, no); \textit{Color of shoes} (dark, light); \textit{Color of upper-body clothes} (black, white, red, purple,
gray, yellow, blue, green, blown); \textit{Color of lower-body clothes} (black, white, red, purple, gray, yellow, blue, green, blown).
Some representative samples with attribute label at the identity level are also demonstrated in Fig.~\ref{fig11}.
Intuitively, there exists a potential direction that our \textit{FineGPR} can also be applied in other surveillance task (\textit{e.g.} pedestrian attribute recognition) for fine-grained analysis.

\begin{figure}[!t]
\centerline{\includegraphics[width=1.00\linewidth]{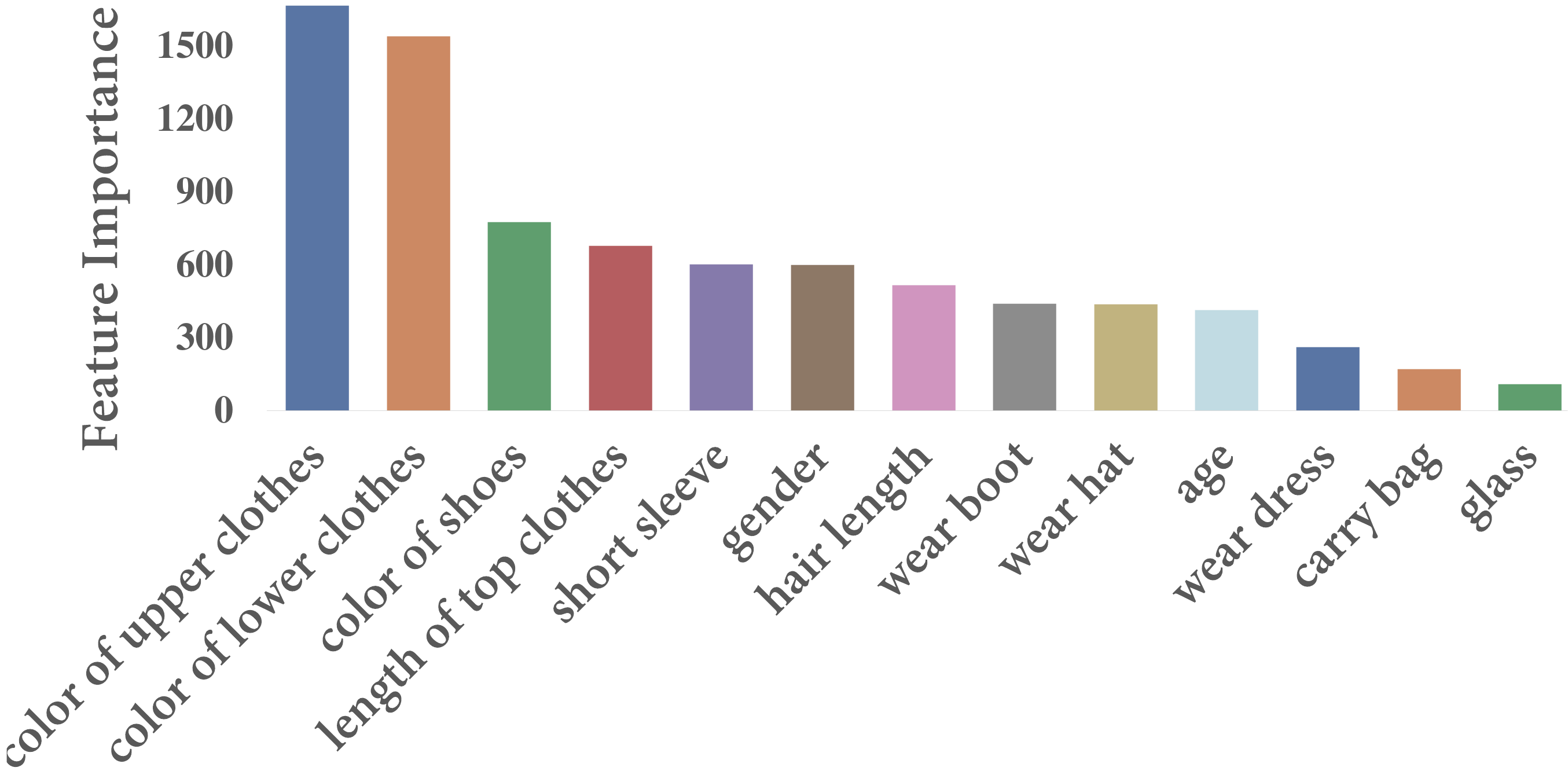}}
\caption{Feature importance at Identity-level with Feature Importance Score (higher is more important) on Market-1501.}
\label{fig3}
\end{figure}

\section{More Experiment Results}

\subsection{Analysis of ID-level Attribute Importance}
Based on the end-to-end AOST system, we evaluate the impacts of different attributes in a fine-grained manner by the boosting system
XGBoost in terms of \textit{gain}$\footnote[1]{\textcolor{black}{https://xgboost.readthedocs.io/en/latest/tutorials/model.html}}$ (feature importance score). As illustrated in Fig.~\ref{fig3}, it can be easily observed that \textbf{Colors of upper-body clothes} and \textbf{Colors of lower-body clothes} account for the largest proportion when testing in real-world domain. This observation is consistent with our intuition because
the color of upper-body /lower-body clothes contain more detailed and discriminative information,  a model trained with discriminative cloth appearances is good at recognizing the pedestrians under different orientations.

We also provide feature distance distribution  $\mathcal{D}_{\text {total}}$ by our attribute optimization on Market-1501, which help us better understand the distance distribution based on our AOST with fine-grained \textit{FineGPR} dataset. As shown in Fig.~\ref{fig8}, those attributes whose value below \textcolor{red}{horizontal line} can be regarded as optimal items, which is more closer to real-world scenarios in re-ID task.

\subsection{Visualization of Style Transfer Results}
Fig.~\ref{fig4} demonstrates the translated images in high-resolution settings. To be more specific, we train the SPGAN with MSCO dataset instead of original low-resolution images (\textit{e.g.} Market-1501). Consequently, when compared with original synthetic data \textit{FineGPR}, our method can generate more consistent and photo-realistic images in a even better fashion, which proves that using a unified real-world dataset with high-resolution can learn more discriminative embeddings and bring great benefits to the GAN models. Since previous PersonX, Unreal and RandPerson datasets do not provide fine-grained attribute annotations, which fail to satisfy the need for AOST in generalizable re-ID task. To solve this problem, we instead randomly sample 124,200 images from these datasets individually, and then perform style transfer. The detailed analysis can be listed as following:

\begin{figure}[t]
\begin{minipage}[t]{0.5\linewidth}
\centering
\includegraphics[width=1.8in]{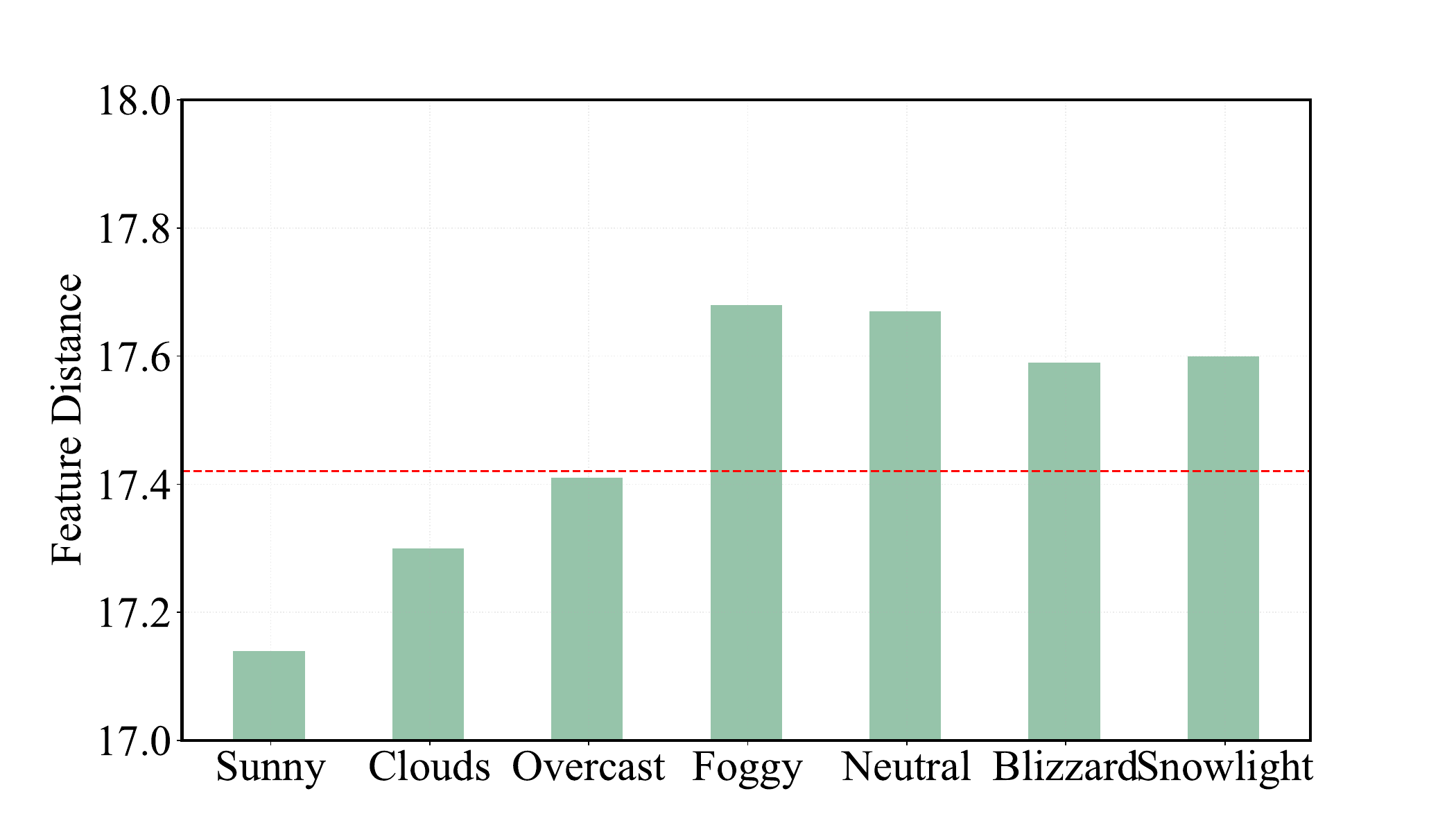}
\centerline{(a) Weather}
\label{fig8a}
\end{minipage}%
\begin{minipage}[t]{0.5\linewidth}
\centering
\includegraphics[width=1.8in]{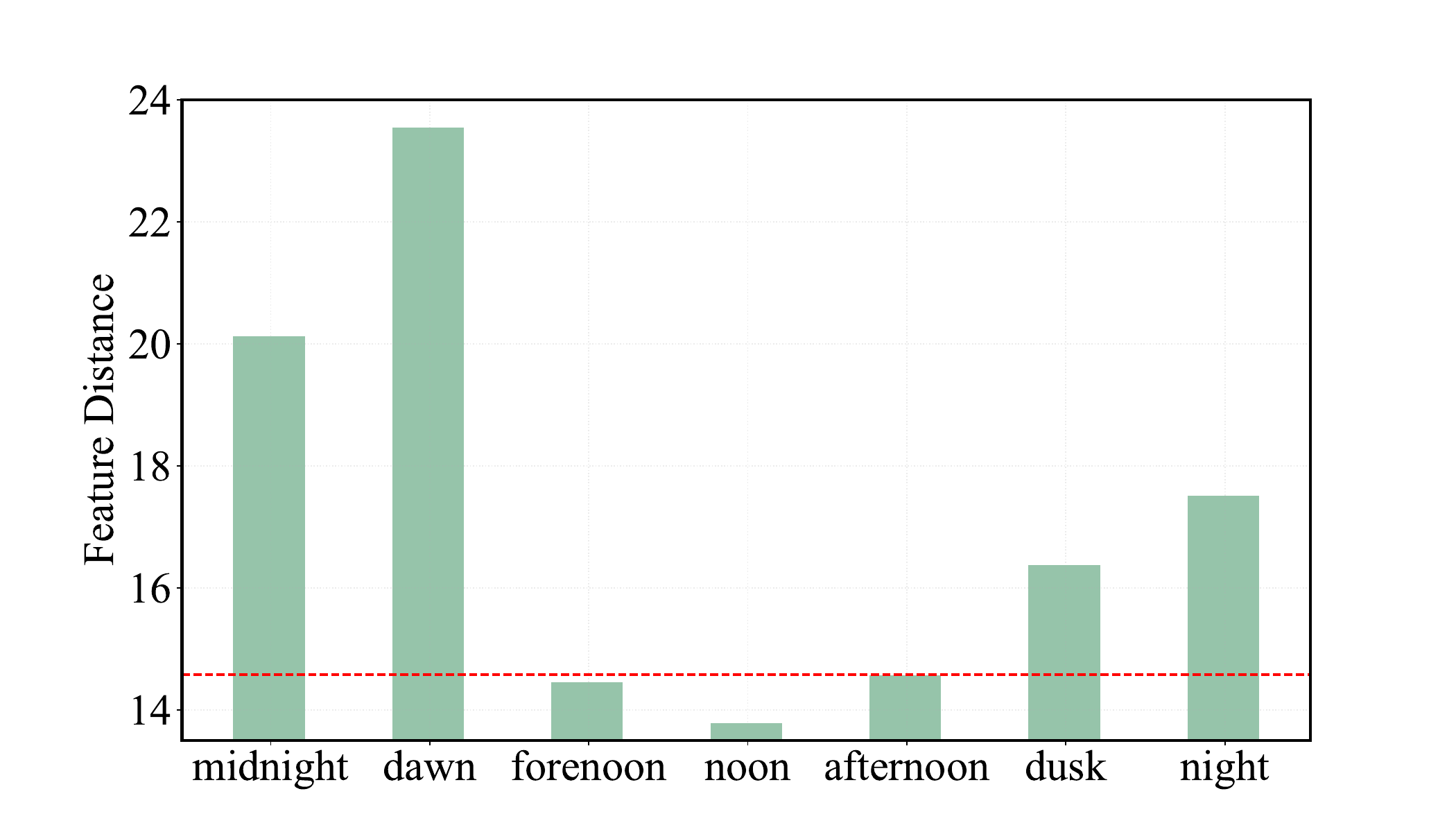}
\centerline{(b) Illumination}
\label{fig8b}
\end{minipage}
\caption{Feature distances $\mathcal{D}_{\text {total}}$ in terms of \textbf{weather} and \textbf{illumination} distribution respectively on Market-1501. Attributes whose value below \textcolor{red}{horizontal line} can be regarded as optimal items.}
\label{fig8}
\end{figure}

From a qualitative point of view, as shown in Fig.~\ref{fig10}, Fig.~\ref{fig15} and Fig.~\ref{fig6}, it can be obviously observed that
the  style transfer results of PersonX+R+ST, Unreal+R+ST and RandPerson+R+ST tend to loss some texture feature and produce some abnormal color (Marked with \textcolor{red}{red} rectangle). We argue that this is due to the samples in PersonX, Unreal and
RandPerson are low-resolution, and GANs tend to have semantic shifts and create some noises inevitably, which seriously
degrades the performance of re-ID model.

\begin{table*}[htbp]
  \centering
  \footnotesize
  \setlength{\tabcolsep}{2.3mm}{
    \begin{tabular}{l|c|c|ccc|ccc|ccc}
    \Xhline{0.8pt}
    \multicolumn{3}{c|}{Testing set $\rightarrow$} & \multicolumn{3}{c|}{Market-1501} & \multicolumn{3}{c|}{MSMT17} & \multicolumn{3}{c}{CUHK03} \bigstrut\\
    \hline\hline
    \multicolumn{1}{l|}{ Training set $\downarrow$} & \multicolumn{1}{l|}{Bboxes} & viewpoint & Rank-1 & Rank-5 & mAP   & Rank-1 & Rank-5 & mAP   & Rank-1 & Rank-5 & mAP \bigstrut\\
    \hline
    \multirow{3}[2]{*}{\textit{FineGPR}} & 225400 & front-view & \textbf{47.2}  & \textbf{65.0}  & \textbf{22.9}  & \textbf{10.1}  & \textbf{18.4}  & \textbf{2.7}  & \textbf{10.8}  & \textbf{21.8}  & \textbf{9.3}  \bigstrut[t]\\
          & 225400 & side-view & 46.8  & 64.2  & 22.5  & 8.8  & 16.9  & 2.3  & 8.1   & 17.9  & 7.7  \\
          & 225400 & random-view & 46.6  & 63.2  & 21.9  & 9.1  & 16.9  & 2.4  & 9.6   & 19.4  & 8.7  \\
    \Xhline{0.8pt}
    \end{tabular}}%
    \caption{Evaluation performance (\%) of baseline model on Market-1501, MSMT17 and CUHK03 respectively in terms of
  viewpoint, \textit{e.g.}, front-view, side-view and random-view.}
  \label{tab1}%
\end{table*}%
\begin{table*}[htbp]
  \centering
  \footnotesize
  \setlength{\tabcolsep}{2.3mm}{
    \begin{tabular}{l|c|ccc|ccc|ccc}
    \Xhline{0.8pt}
    \multicolumn{2}{c|}{Testing set $\rightarrow$} & \multicolumn{3}{c|}{Market-1501} & \multicolumn{3}{c|}{MSMT17} & \multicolumn{3}{c}{CUHK03} \bigstrut\\
    \hline\hline
    Training set $\downarrow$ & Datasets Fusion  & Rank-1 & Rank-5 & mAP   & Rank-1 & Rank-5 & mAP   & Rank-1 & Rank-5 & mAP \bigstrut\\
    \hline
    Market & $\times$     & \textit{\underline{92.7}}  & \textit{\underline{97.9}}  & \textit{\underline{81.4}}  & 6.0  & 11.2  & 1.9  & 5.3   & 12.4  & 6.2  \\
    MSMT17  & $\times$     & 50.2  & 67.7  & 25.7  & \textit{\underline{75.7}}  & \textit{\underline{86.9}}  & \textit{\underline{51.5}}  & 9.9   & 20.4  & 10.7  \\
    CUHK03 & $\times$     & 36.6  & 53.9  & 16.6  & 4.6  & 9.1  & 1.3   & \textit{\underline{43.6}}  & \textit{\underline{62.9}}  & \textit{\underline{41.5}}  \\
    \hline
    \textit{FineGPR}+R & $\times$     & 39.9  & 57.1  & 18.3  & 4.6  & 9.7  & 1.4  & 5.9   & 15.4  & 5.4  \\
    \textit{FineGPR}+R+Market & \checkmark       & \textit{\underline{92.6}}    & \textit{\underline{97.4}}    & \textit{\underline{82.4}}    & 8.3  & 15.4  & 2.9  & 14.1  & 25.8  & 13.1  \\
    \textit{FineGPR}+R+MSMT17 & \checkmark       & 56.0  & 71.4  & 30.3  & \textit{\underline{71.2}}    & \textit{\underline{83.3}}    & \textit{\underline{47.0}}    & 14.2  & 27.3  & 14.5  \\
    \textit{FineGPR}+R+CUHK03 & \checkmark       & 42.4  & 60.4  & 20.7  & 7.0  & 13.5  & 2.2  & \textit{\underline{37.9}}    & \textit{\underline{57.3}}    & \textit{\underline{36.4}} \\
    \Xhline{0.8pt}
    \end{tabular}}%
    \caption{Evaluation performance (\%) of baseline model trained on various datasets and tested on Market-1501,
  MSMT17 and CUHK03 respectively. ``R" indicates random sampling.
  \textit{\underline{Underline}} denodes supervised learning.}
  \label{tab4}%
\end{table*}%

\begin{figure}[!t]
\centerline{\includegraphics[width=1.0\linewidth]{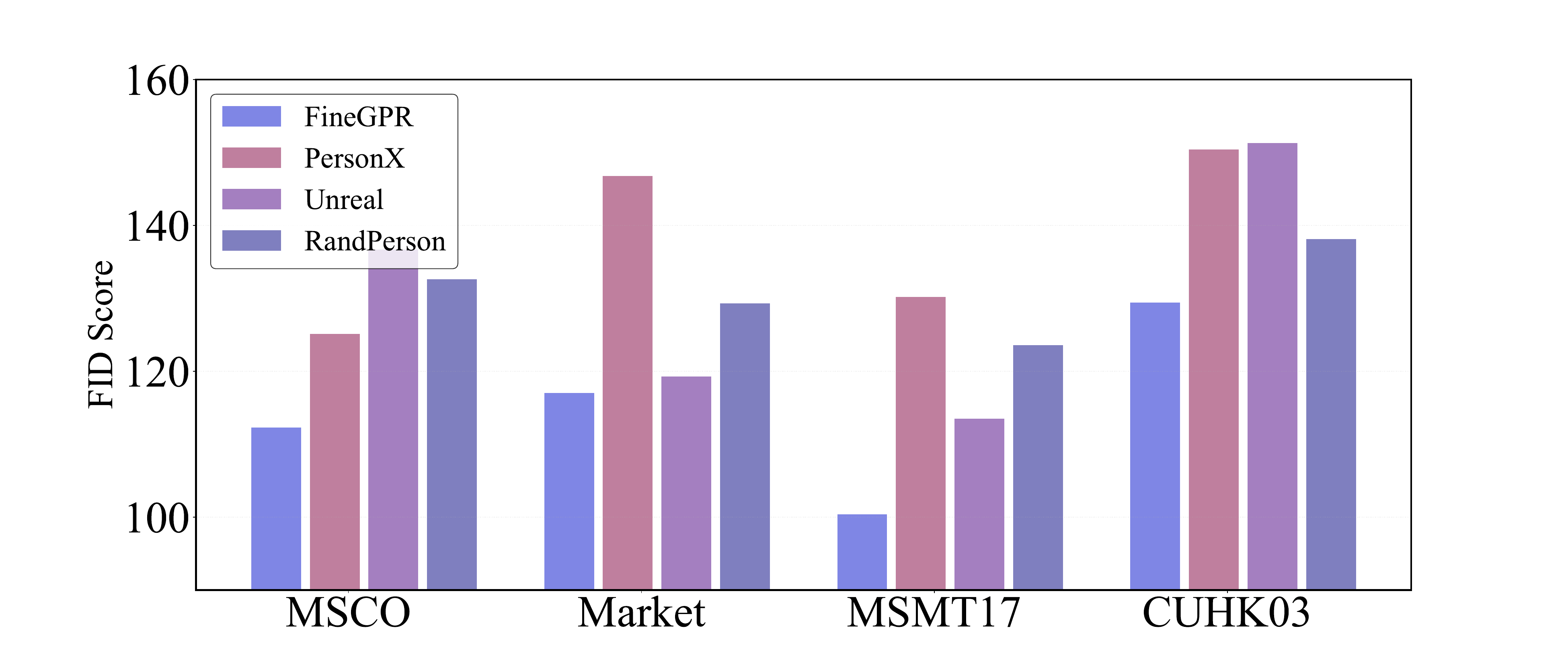}}
\caption{Comparison of FID (lower is better) to evaluate the realism of style transferred images of \textit{FineGPR}, PersonX, Unreal and RandPerson respectivey when comparing with real images, \textit{e.g.}, MSCO, Market-1501, MSMT17 and CUHK03 individually.}
\label{fig12}
\end{figure}

More importantly, from a quantitative point of view, we also provide a quantitative analysis on the quality of transferred images of \textit{FineGPR}, PersonX, Unreal and RandPerson.
As shown in Fig.~\ref{fig12}, when compared to the real-world images in terms of realism, \textit{e.g.}, MSCO, Market-1501, MSMT17 and CUHK03, the transferred images of \textit{FineGPR} always have lowest FID score than transferred samples of PersonX, Unreal and RandPerson,  this significantly validates the priority of our transferred \textit{FineGPR} dataset.

\subsection{Visualization of Attribute Optimization Results}
\begin{figure}[!t]
\centerline{\includegraphics[width=0.90\linewidth]{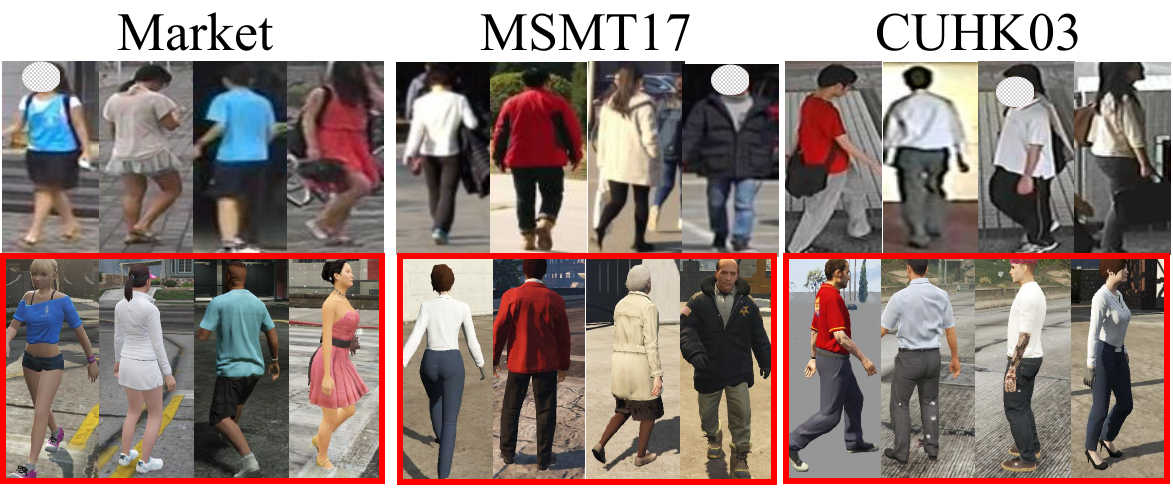}}
\caption{Some visual examples selected by the attribute optimization on different datasets.
Each group surrounded by the \textcolor{red}{red box} is the similar pair mined by our model.}
\label{fig16}
\end{figure}

To further validate the effectiveness of our attribute optimization method, we present
some visual exemplars qualitatively \textit{w.r.t} different real target datasets. As shown in Fig.~\ref{fig16}, it can be obviously observed
that the selected samples from \textit{FineGPR} are more similar to the real target samples in terms of style and content representation, demonstrating the effectiveness of our optimization method.

\subsection{Impact of Viewpoint}
Based on the \textit{FineGPR}, we further evaluate the impact of viewpoint on generalizable person re-ID quantitatively. The experiment is based on the re-ID baseline in the main manuscript.
We note that each subset contains 225,400 images with 1,150 identities of 4 kinds of different angles.
The detailed viewpoint settings are designed as following:

$\bullet$ Experimental group 1. \textbf{Random-view} Setting. This subset includes samples randomly selected from \textit{FineGPR}
with angles of $100^{\circ}$, $150^{\circ}$, $190^{\circ}$ and $350^{\circ}$.

$\bullet$ Experimental group 2. \textbf{Side-view} Setting. This subset includes samples from \textit{FineGPR} with angles of $40^{\circ}$, $130^{\circ}$, $220^{\circ}$ and $310^{\circ}$.

$\bullet$ Experimental group 3. \textbf{Front-view} Setting. This subset includes samples from \textit{FineGPR} with angles of $0^{\circ}$, $90^{\circ}$, $180^{\circ}$ and $270^{\circ}$.

According to the Table~\ref{tab1}, the major observation is that there seems to exist obvious correlation between viewpoint distribution and performance results. To be more specific, using samples with \textbf{Front-view} setting can always bring significant performance gains to the re-ID
system, no matter which dataset is adopted as testing-set, (\textit{e.g.} Market-1501, MSMT17 and CUHK03). This observation is intuitive because the samples with front-view contain more discriminative information and clothes appearance of human body, which help to learn a discriminative re-ID model. Consequently, the performance can be greatly improved when identifying pedestrians under other different orientations.

\begin{figure}[!t]
\centerline{\includegraphics[width=0.9\linewidth]{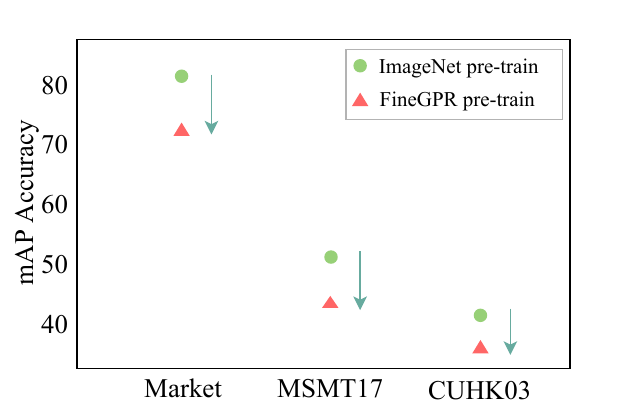}}
\caption{Performance comparison with ImageNet and \textit{FineGPR} dataset respectively for pre-training in supervised re-ID
tasks. Best viewed in high resolution.}
\label{fig13}
\end{figure}

\subsection{Evaluation of Dateset Fusion}
To further verify the superiority of our proposed \textit{FineGPR}, we perform another round of experiments with dataset fusion strategy. To be more specific, we combine our synthetic dataset \textit{FineGPR} with single real-world datasets during the experiment, \textit{e.g.}, Market-1501, MSMT17 and CUHK03. In particular, we randomly select 124,200 images from our synthetic dataset (Denoted as \textit{FineGPR+R}). In this case, as reported in Table~\ref{tab4}, there are two observations which can be made.

First, when compared to training with single real-world dataset, dataset fusion strategy can always boost the generalizable re-ID performance into a satisfactory level. For example, as illustrated in Table~\ref{tab4}, we can achieve a rank-1 accuracy of 56.0\% and 42.4\% on Market-1501 when trained with \textit{FineGPR}+R+MSMT17 and \textit{FineGPR}+R+CUHK03 respectively, outperforming the single-dataset training (MSMT17 or CUHK03) by \textbf{+5.8\%} and \textbf{+5.8\%}, respectively.

Second and importantly, compared to only using \textit{FineGPR}+R for training, as can be observed in Table~\ref{tab4}, dataset fusion strategy can significantly boost the performance from 4.6\% to 8.3\% and 7.0\% on MSMT17 when combining \textit{FineGPR}+R with Market-1501 and CUHK03 respectively. It can be concluded that combining synthetic data with real-world data for training can bridge the domain gap between synthetic and real-world datasets and help to learn a robust and discriminative model.

\subsection{Pre-training}
\textit{Does FineGPR can replace ImageNet for pre-training?} To find the answer to this question, we use different pre-trained models (ImageNet pre-train and \textit{FineGPR} pre-train) on the re-ID baseline in a traditional pre-train manner. As shown in Fig.~\ref{fig13}, it can be seen that, pre-training with \textit{FineGPR} achieves slightly inferior performance than ImageNet pre-training, but still acceptable.
One potential reason is the large domain gap between real and synthetic images. Consequently, we suspect that this fine-grained attribute dataset
is not applicable to the traditional pre-training of coarse-grained retrieval tasks in some degree.

Surprisingly, recent research work by Xiang \textit{et al.}~\cite{xiang2021vtbr} has given re-ID community some insights on pre-training, which demonstrated that a pure semantic-based pre-training approach VTBR$\footnote[2]{\textcolor{black}{The VTBR adopts another total different re-ID backbone.}}$, by leveraging the fine-grained attributes of \textit{FineGPR}, can outperform the traditional ImageNet pretraing by a clear margin (\textit{e.g.}, mAP \textcolor[rgb]{1.00,0.39,0.09}{\textbf{85.3\%}} vs. \textcolor[rgb]{0.20,0.40,0.80}{\textbf{85.0\%}} on Market-1501 in a supervised manner), so \textit{FineGPR} can still reveal its potential in re-ID pretraining. Further details related to pretraining may be found in~\cite{xiang2021vtbr}.

\section{Extendibility}
\subsection{More Tasks}
Accompanied with our \textit{FineGPR}, we also provide some \textbf{human body masks} and \textbf{keypoint locations} of all characters during the annotation. The standard skeleton provided among with our \textit{FineGPR} has 17 keypoints, which is obtained by AlphaPose~\cite{fang2017rmpe}, and segmentation for human mask is extracted by PSPNet~\cite{zhao2017pyramid}. More pose estimation results and segmentation results are shown in Fig.~\ref{fig5}. For example, we have adopted human body masks to reproduce style transfer results with PTGAN. To sum up, we hope that our synthetic dataset \textit{FineGPR} can not only contribute a lot to the development of generalizable person re-ID, but also advance the research of other computer vision tasks, such as human part segmentation and pose estimation.

\subsection{More Identities}
According to the Table 2 in the main manuscript, we observe that our synthetic \textit{FineGPR} can achieve a satisfactory performance in most cases for generalizable re-ID task. However, it is still not as good as RandPerson where 132,145 person images of 8,000 identities are adopted for training.
We suspect that the success of RandPerson can be largely attributed to the huge number of identities, which can greatly
enhance the discrimination capability of backbone network.
Note that for fairness, we manually construct a ID-enhanced version of \textit{FineGPR} dataset (named as \textit{FineGPR\_V2}), which contains 137,808 samples of 6506 identities$\footnote[2]{\textcolor{black}{The ID-enhanced version of \textit{FineGPR\_V2} only provides annotations in terms of identity and viewpoint.}}$. As illustrated in Table~\ref{tab6},
it can be easily observed that the \textit{FineGPR\_V2} can achieve competitive performance with RandPerson when Identities are highly promoted, which again verified the attribute importance of Identity. (Consistent with the conclusion in the Section 5.4 of the main manuscript).

\begin{table}[!t]
  \centering
  \footnotesize
  \setlength{\tabcolsep}{1.8mm}{
    \begin{tabular}{l|cc|cc|cc}
    \Xhline{0.8pt}
    Testing set $\rightarrow$ & \multicolumn{2}{c|}{Market-1501} & \multicolumn{2}{c|}{MSMT17} & \multicolumn{2}{c}{CUHK03} \bigstrut\\
    \hline\hline
    Training set $\downarrow$   & Rank-1 & mAP   & Rank-1 & mAP   & Rank-1 & mAP \bigstrut\\
    \hline
    RandPerson      & 55.6   & 28.8  & 20.1   & 6.3  & 13.4   & 10.8  \\
    \textit{FineGPR\_V2}     & 53.7   & 26.6  & 18.8   & 5.4  & 12.5   & 9.8  \\
    \Xhline{0.8pt}
    \end{tabular}}%
    \caption{Performance comparison between RandPerson and our enhanced \textit{FineGPR\_V2} dataset.}
  \label{tab6}%
\end{table}%

\section{Limitations}
\begin{figure}[!t]
\centerline{\includegraphics[width=0.95\linewidth]{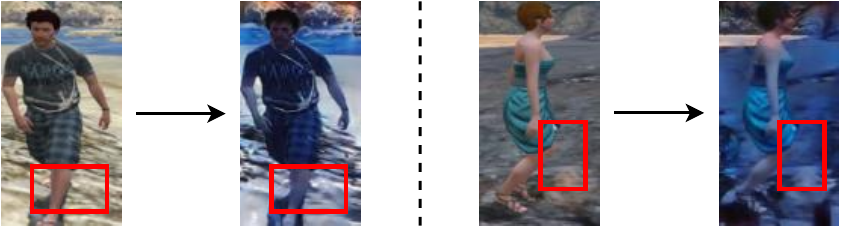}}
\caption{Some failure cases in the style transfer.}
\label{fig17}
\end{figure}

\begin{figure*}[htbp]
\centerline{\includegraphics[width=1.0\linewidth]{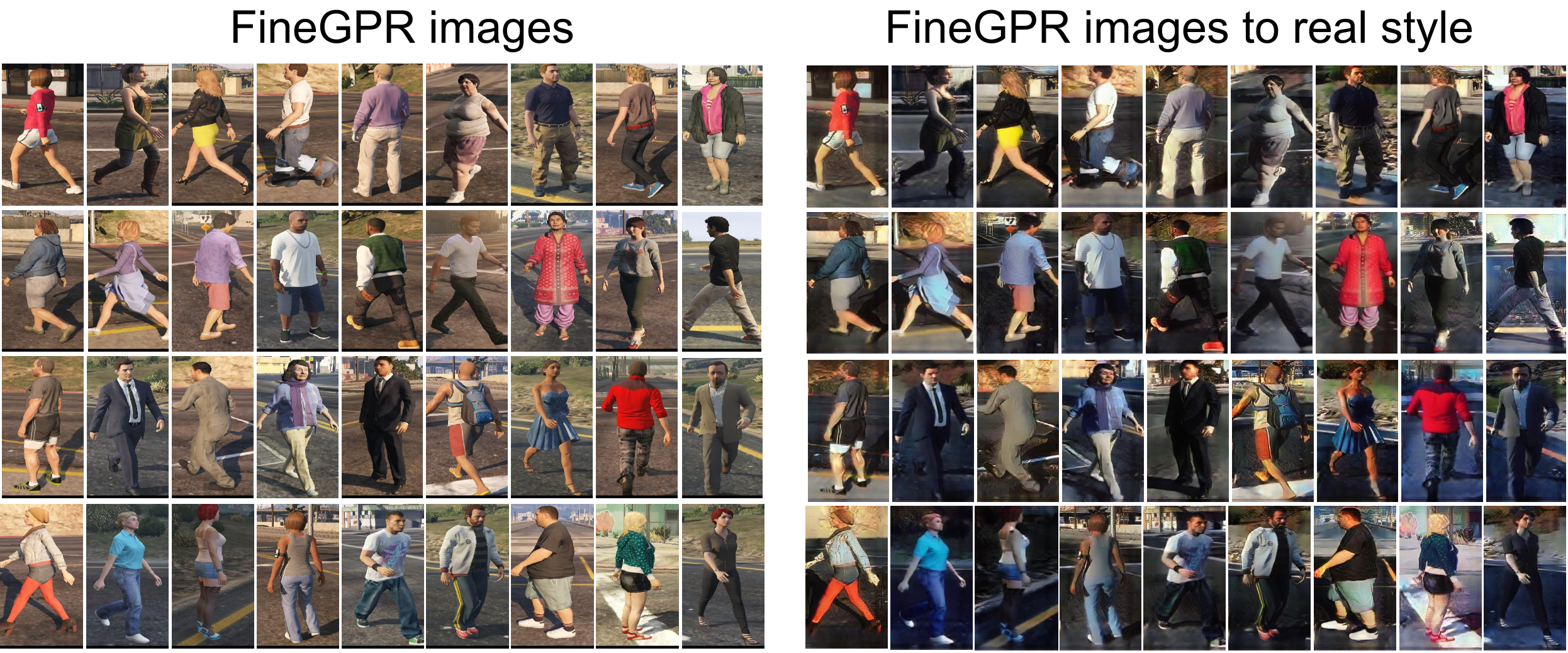}}
\caption{Some visual samples of transferred person images from synthetic \textit{FineGPR} dataset to photo-realistic style by SPGAN. (\textit{FineGPR}$\rightarrow$MSCO). It can be easily observed that there exist obvious differences between original synthetic \textit{FineGPR} and our transferred results with photo-realistic style.}
\label{fig4}
\end{figure*}

\begin{figure*}[htbp]
\centerline{\includegraphics[width=1.0\linewidth]{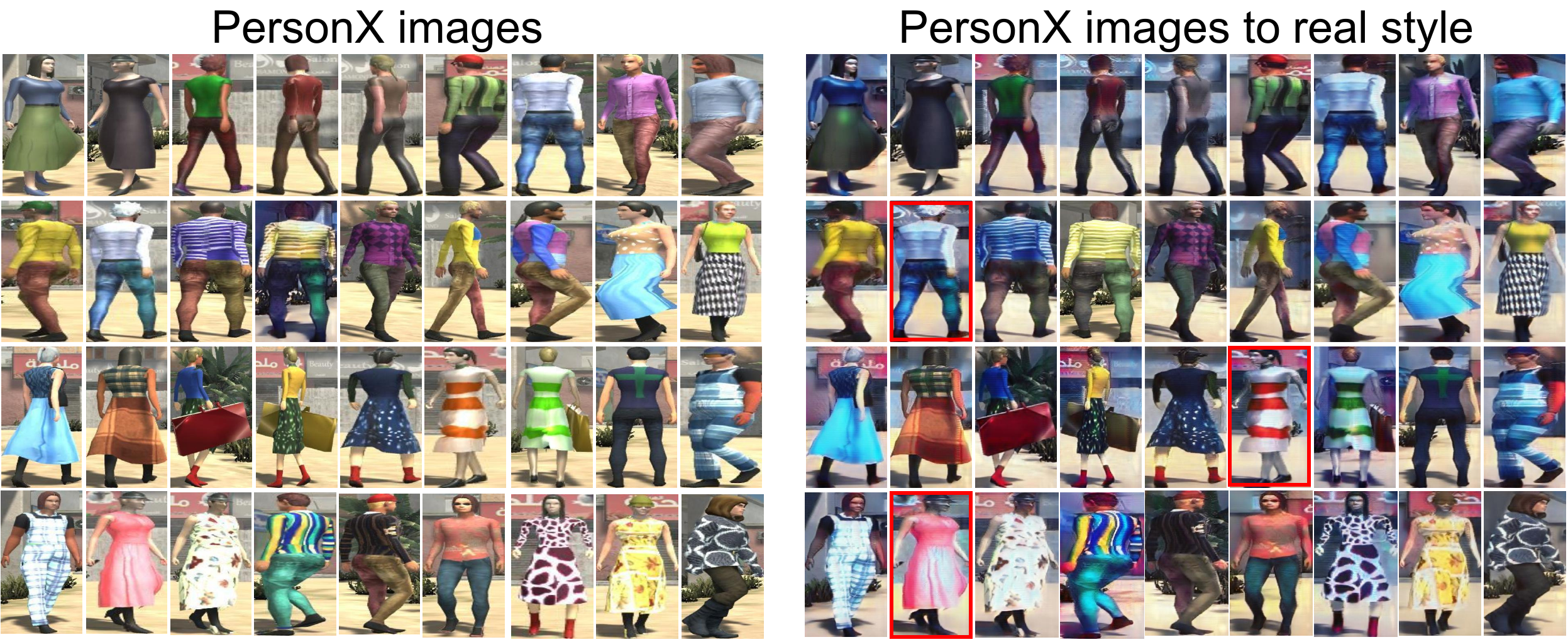}}
\caption{Some visual samples of transferred person images from synthetic PersonX dataset to photo-realistic style by SPGAN (PersonX$\rightarrow$MSCO).
Some cases tend to produce some abnormal color (Marked with \textcolor{red}{red} rectangle). Please zoom in for better observation.
}
\label{fig10}
\end{figure*}

\begin{figure*}[htbp]
\centerline{\includegraphics[width=1.0\linewidth]{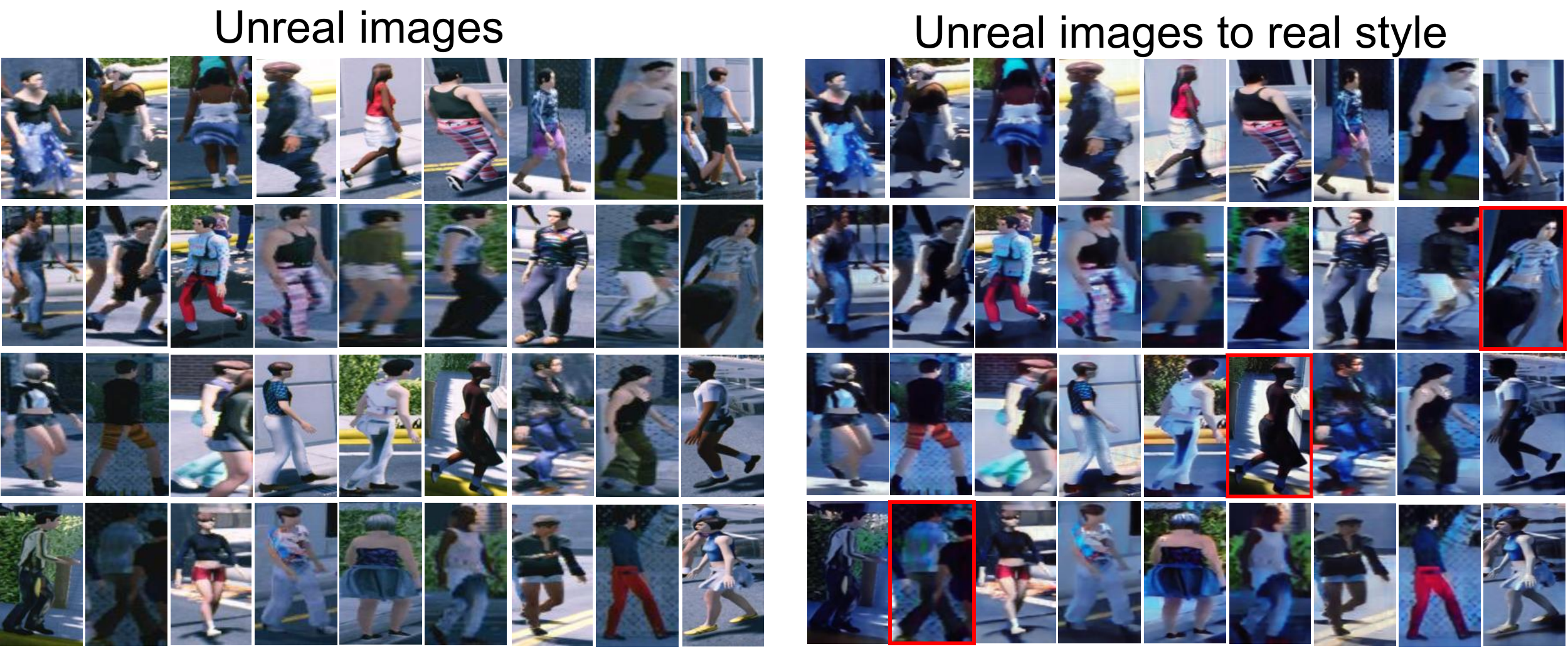}}
\caption{Some visual samples of transferred person images from Unreal dataset to photo-realistic style by SPGAN (Unreal$\rightarrow$MSCO).
Some cases tend to produce some abnormal color (Marked with \textcolor{red}{red} rectangle). Please zoom in for better observation.
}
\label{fig15}
\end{figure*}

\begin{figure*}[htbp]
\centerline{\includegraphics[width=1.0\linewidth]{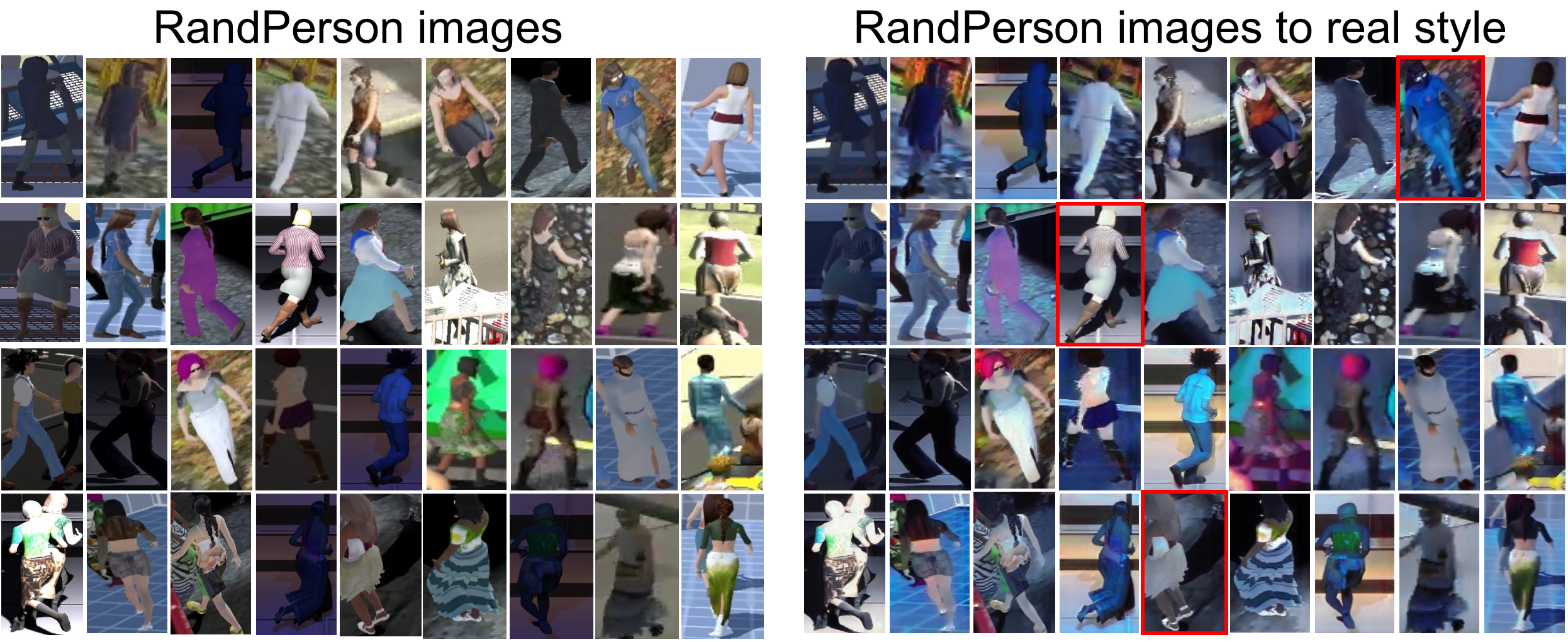}}
\caption{Some visual samples of transferred person images from RandPerson dataset to photo-realistic style by SPGAN (RandPerson$\rightarrow$MSCO).
Some cases tend to produce some abnormal color (Marked with \textcolor{red}{red} rectangle). Please zoom in for better observation.
}
\label{fig6}
\end{figure*}

\begin{figure*}[htbp]
\centerline{\includegraphics[width=1.0\linewidth]{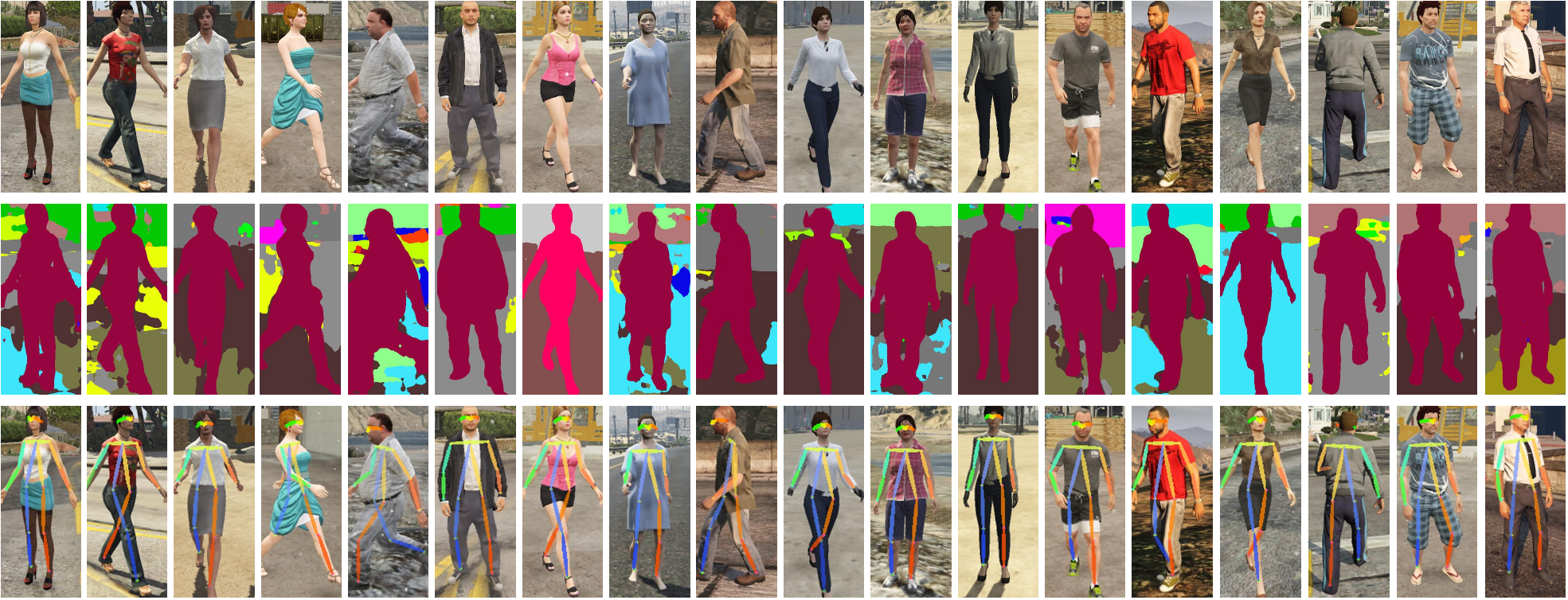}}
\caption{Some visual results with human body segmentation (\textbf{Middle}) and pose estimation (\textbf{Bottom}) based on our \textit{FineGPR} dataset.}
\label{fig5}
\end{figure*}

\begin{figure*}[htbp]
\centerline{\includegraphics[width=0.95\linewidth]{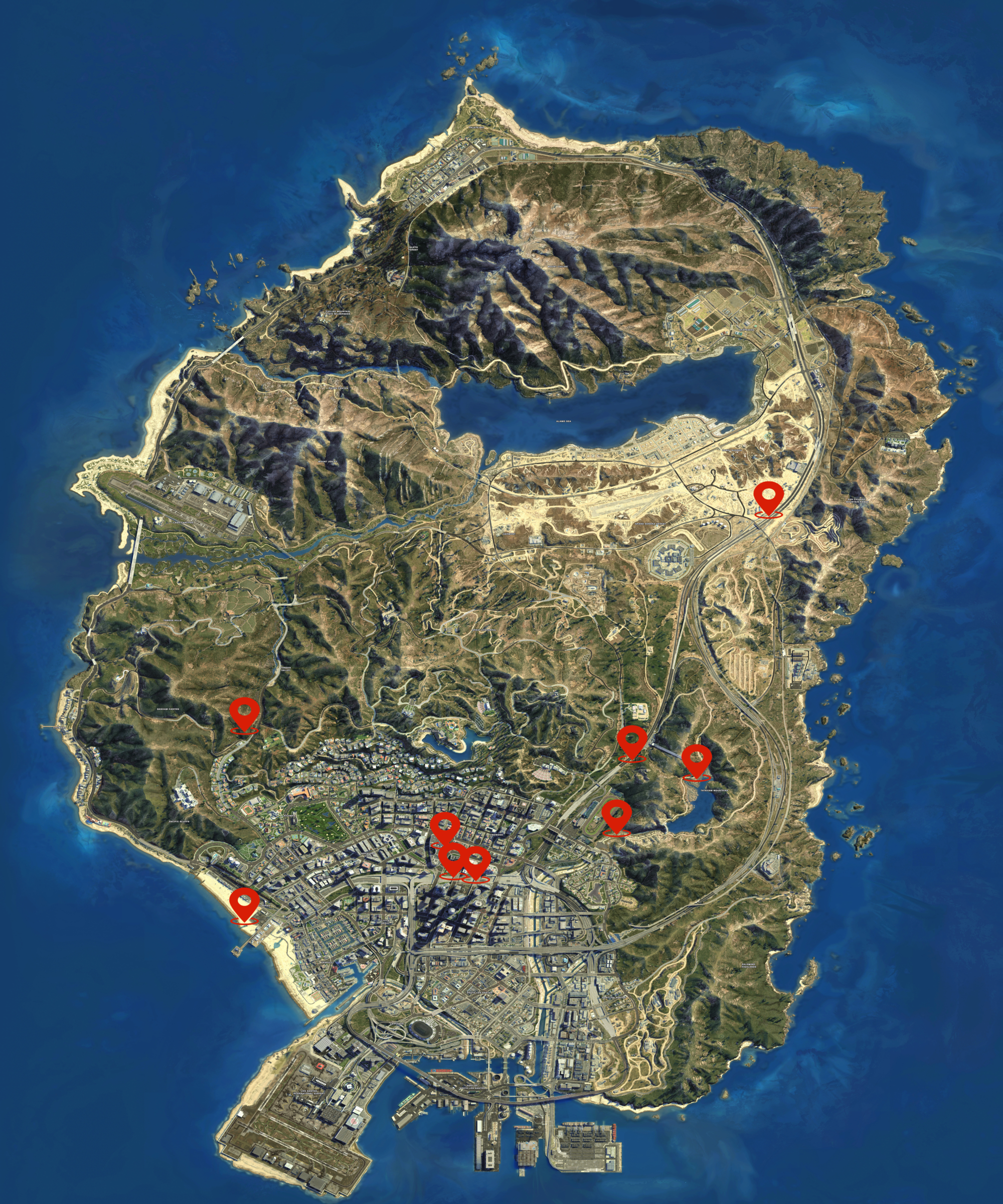}}
\caption{The position of each location of 9 backgrounds from our proposed \textit{FineGPR} dataset in GTA5 world.}
\label{fig7}
\end{figure*}

The proposed method in this paper can dynamically mine some critical pedestrian samples which is closer to the specific domain, then perform style transfer. Despite its promising performance on person re-ID, we note that there are several limitations in AOST, including (1) the premise of our AOST is that the attribute distribution of target domain are fixed, this means the sample selection process might be repeated when target domain is gradually updated; (2) the GAN model will generate semantic shifts in some extremely cases, as shown in Fig.~\ref{fig17}, which may bring negative impacts on downstream tasks. We believe addressing these challenges are promising directions of our work for future research.

\begin{table*}[!t]
  \centering
  \setlength{\tabcolsep}{2.2mm}{
    \begin{tabular}{|l|c|c|}
    \hline
    \multicolumn{1}{|l|}{Attribute} & Category name & \multicolumn{1}{l|}{Label ID} \bigstrut\\
    \hline
    \multirow{2}[4]{*}{Gender} & male  & 01  \\
\cline{2-3}          & female & 02 \\
    \hline
    \multirow{2}[4]{*}{Hair length} & long   & 01 \\
\cline{2-3}          & short    & 02 \\
    \hline
    \multirow{3}[6]{*}{Age} & teenager   & 01 \\
\cline{2-3}          & adult & 02 \\
\cline{2-3}          & older & 03 \\
    \hline
    \multirow{2}[4]{*}{Wear glass} & yes   & 01 \\
\cline{2-3}          & no    & 02 \\
    \hline
    \multirow{2}[4]{*}{Wear short sleeve} & yes   & 01 \\
\cline{2-3}          & no    & 02 \\
    \hline
    \multirow{2}[4]{*}{Length of top clothes} & long  & 01 \\
\cline{2-3}          & short & 02 \\
    \hline
    \multirow{2}[4]{*}{Wear dress} & yes   & 01 \\
\cline{2-3}          & no    & 02 \\
    \hline
    \multirow{2}[4]{*}{Wear boot} & yes   & 01 \\
\cline{2-3}          & no    & 02 \\
    \hline
    \multirow{2}[4]{*}{Wear hat} & yes   & 01 \\
\cline{2-3}          & no    & 02 \\
    \hline
    \multirow{2}[4]{*}{Carry bag} & yes   & 01 \\
\cline{2-3}          & no    & 02 \\
    \hline
    \multirow{2}[4]{*}{Color of shoes} & dark  & 01 \\
\cline{2-3}          & light & 02 \\
    \hline
    \multirow{6}[16]{*}{Colors of upper-body clothes} & black & 01 \\
\cline{2-3}          & white & 02 \\
\cline{2-3}          & red & 03 \\
\cline{2-3}          & purple & 04 \\
\cline{2-3}          & gray & 05 \\
\cline{2-3}          & yellow & 06 \\
\cline{2-3}          & blue & 07 \\
\cline{2-3}          & green & 08 \\
\cline{2-3}          & brown & 09 \\
    \hline
    \multirow{6}[16]{*}{Colors of lower-body clothes} & black & 01 \\
\cline{2-3}          & white & 02 \\
\cline{2-3}          & red & 03 \\
\cline{2-3}          & purple & 04 \\
\cline{2-3}          & gray & 05 \\
\cline{2-3}          & yellow & 06 \\
\cline{2-3}          & blue & 07 \\
\cline{2-3}          & green & 08 \\
\cline{2-3}          & brown & 09 \\
    \hline
    \end{tabular}}%
\caption{List of all the ID-level attribute annotations in \textit{FineGPR}. The relative class id is also provided.}
  \label{tab5}%
\end{table*}%

{\small
\bibliographystyle{ieee_fullname}
\bibliography{egbib}
}